\documentclass[runningheads]{llncs}
\usepackage{graphicx}

\usepackage{tikz}
\usepackage{comment}
\usepackage{amsmath,amssymb} 
\usepackage{color}
\usepackage{subcaption}

\usepackage{graphicx}
\usepackage{amsmath}
\usepackage{amssymb}
\usepackage{booktabs}
\usepackage{float}

\usepackage{multirow}
\usepackage{xcolor}
\usepackage{wrapfig}
\usepackage{verbatim}
\usepackage{caption}
\usepackage{ulem}
\usepackage{placeins}
\usepackage{ragged2e}
\usepackage{diagbox}
\usepackage{enumitem}

\usepackage{epsfig}
\usepackage{textcomp}
\usepackage{soul}

\usepackage{makecell}
\usepackage{arydshln}
\usepackage[pagebackref,breaklinks,colorlinks]{hyperref}

\useunder{\uline}{\ul}{}
\newcommand{\ieours}{\textit{i}.\textit{e}., }
\newcommand{\egours}{\textit{e}.\textit{g}., }

\newcommand{\nickname}{SQN}
\newcommand{\nicknamenew}{Ours}

\usepackage[accsupp]{axessibility}  

\begin{document}
\pagestyle{headings}
\mainmatter

\title{\nickname{}: Weakly-Supervised Semantic Segmentation of Large-Scale 3D Point Clouds}

\titlerunning{Semantic Query Network}
%
\author{Qingyong Hu\textsuperscript{1}, Bo Yang\textsuperscript{2\thanks{Corresponding author}}, Guangchi Fang\textsuperscript{3}, Yulan Guo\textsuperscript{3}, Ales Leonardis\textsuperscript{4},\\ Niki Trigoni\textsuperscript{1}, Andrew Markham\textsuperscript{1} \\}
\authorrunning{Q. Hu et al.}
%
\institute{University of Oxford \and The Hong Kong Polytechnic University \and Sun Yat-sen University \and Huawei Noah’s Ark Lab \\
\email{\{qingyong.hu,andrew.markham\}@cs.ox.ac.uk, bo.yang@polyu.edu.hk}\\
}
\maketitle

\begin{abstract}
   Labelling point clouds fully is highly time-consuming and costly. As larger point cloud datasets with billions of points become more common, we ask whether the full annotation is even necessary, demonstrating that existing baselines designed under a fully annotated assumption only degrade slightly even when faced with 1\% random point annotations. However, beyond this point, \textit{e.g.,} at 0.1\% annotations, segmentation accuracy is unacceptably low. We observe that, as point clouds are samples of the 3D world, the distribution of points in a local neighbourhood is relatively homogeneous, exhibiting strong semantic similarity. Motivated by this, we propose a new weak supervision method to implicitly augment highly sparse supervision signals. Extensive experiments demonstrate the proposed Semantic Query Network (\nickname{}) achieves promising performance on seven large-scale open datasets under weak supervision schemes, while requiring only 0.1\% randomly annotated points for training, greatly reducing annotation cost and effort. The code is available at \url{https://github.com/QingyongHu/SQN}.

\end{abstract}

\section{Introduction}
\label{sec:intro}

Learning precise semantic meanings of large-scale point clouds is crucial for intelligent machines to truly understand complex 3D scenes in the real world. This is a key enabler for autonomous vehicles, augmented reality devices, \textit{etc.}, to quickly interpret the surrounding environment for better navigation and planning.

With the availability of large amounts of labeled 3D data for fully-supervised learning, the task of 3D semantic segmentation has made significant progress in the past four years. Following the seminal works PointNet \cite{qi2017pointnet} and SparseConv \cite{sparse}, a series of sophisticated neural architectures \cite{qi2017pointnet++,li2018pointcnn,4dMinkpwski,hu2019randla,thomas2019kpconv,Point_voxel_cnn,cylinder3d,cheng20212} have been proposed in the literature, greatly improving the accuracy and efficiency of semantic estimation on raw point clouds. The performance of these fully-supervised methods can be further boosted with the aid of self-supervised pre-training representation learning as seen in recent studies \cite{xie2020pointcontrast,liu2020p4contrast,wang2020pre,simclr,zhang2021self,thabet2020self}. The success of these approaches primarily relies on densely annotated per-point semantic labels to train the deep neural networks. However, it is extremely costly to fully annotate 3D point clouds due to the unordered, unstructured, and non-uniform data format (\egours over 1700 person-hours to annotate a typical dataset~\cite{behley2019semantickitti} and around 22.3 minutes for a single indoor scene (5m×5m×2m) \cite{scannet}). In fact, for very large-scale scenarios \egours an entire city, it becomes infeasible to manually label every point in practice.

\begin{figure}[t]
	\begin{center}
		\includegraphics[width=0.75\textwidth]{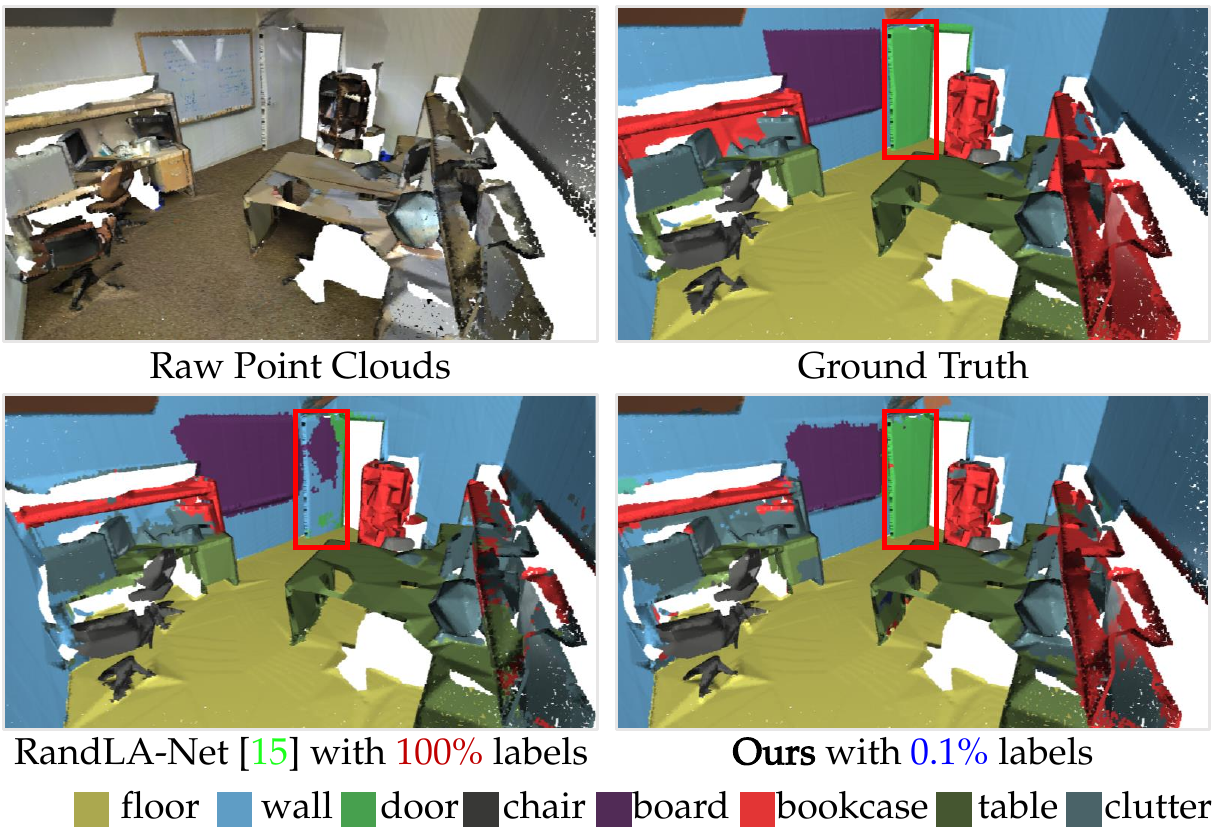}
	\end{center}
    \vspace{-0.2cm}
    \caption{Qualitative results of RandLA-Net \cite{hu2019randla} and our \nickname{} on the S3DIS dataset. Trained with only 0.1\% annotations, \nickname{} achieves comparable or even better results than the fully-supervised RandLA-Net. Red bounding boxes highlight the superior segmentation accuracy of our \nickname{}.}
	\label{fig:teaser}
\end{figure}

Inspired by the success of weakly-supervised learning techniques in 2D images, a few recent works have started to tackle 3D semantic segmentation using fewer point labels to train neural networks. These methods can be generally divided into five categories: 1) Using 2D image labels for training as in \cite{wang2019towards,zhu2021weakly}; 2) Using fewer 3D labels with gradient approximation/supervision propagation/perturbation consistency  \cite{xu2020weakly,zhang2021perturbed,wang2021new,wei2021dense}; 3) Generating pseudo 3D labels from limited indirect annotations \cite{tao2020seggroup,mprm}; 4) Using superpoint annotations from over-segmentation \cite{tao2020seggroup,cheng2021sspc,liu2021one}, and
5) Contrastive pretraining followed by fine-tuning with fewer 3D labels \cite{hou2020exploring,xie2020pointcontrast,Zhang_2021_ICCV}. Although they achieve encouraging results on multiple datasets, there are a number of limitations still to be resolved.

\textbf{Firstly}, existing approaches usually use custom methods to annotate different amounts of data (\egours 10\%/5\%/1\% of raw points or superpoints) for training. It is thus unclear what proportion of raw points should be annotated and how, making fair comparison impossible. \textbf{Secondly}, to fully utilize the sparse annotations, existing weak-labelling pipelines usually involve multiple stages including careful data augmentation, self-pretraining, fine-tuning, and/or post-processing such as the use of dense CRF \cite{densecrf}. As a consequence, it tends to be more difficult to tune the parameters and deploy them in practical applications, compared with the standard end-to-end training scheme. \textbf{Thirdly}, these techniques do not adequately consider the strong local semantic homogeneity of point neighbors in large-scale point clouds, or do so ineffectively, resulting in the limited, yet valuable, annotations being under-exploited.

Motivated by these issues, we propose a new paradigm for weakly-supervised semantic segmentation on large-scale point clouds, addressing the above shortcomings. In particular, we first explore weak-supervision schemes purely based on existing fully-supervised methods, and then introduce an effective approach to learn accurate semantics given extremely limited point annotations.

To explore weak supervision schemes, we take into account two key questions: 1) \textit{whether, and how, do existing fully-supervised methods deteriorate given different amounts of annotated data for training?} 2) \textit{given fewer and fewer labels,  where the weakly supervised regime actually begins?} Fundamentally, by doing so, we aim to explore the limit of current fully-supervised methods. This allows us to draw insights about the use of mature architectures when addressing this challenging task, instead of na\"ively borrowing off-the-shelf techniques developed in 2D images \cite{MT}. Surprisingly, we find that the accuracy of existing fully-supervised baselines drops only slightly when faced with 1\% of random labelled points. However, beyond this point, \egours 0.1\% of the full annotations, the performance degrades rapidly.

With this insight, we propose a novel yet simple \textbf{S}emantic \textbf{Q}uery \textbf{N}etwork, named \textbf{\nickname{}}, for semantic segmentation given as few as 0.1\% labeled points for training. Our \nickname{} firstly encodes the entire raw point cloud into a set of hierarchical latent representations via an existing feature extractor, and then takes an arbitrary 3D point position as input to query a subset of latent representations within a local neighborhood. These queried representations are summarized into a compact vector and then fed into a series of multilayer perceptrons (MLPs) to predict the final semantic label. Fundamentally, our \nickname{} explicitly and effectively considers the semantic similarity between neighboring 3D points, allowing the extremely sparse training signals to be back-propagated to a much wider spatial region, thereby achieving superior performance under weak supervision.

Overall, this paper takes a step to bridge the gap between the highly successful fully-supervised methods to the emerging weakly-supervised schemes, in an attempt to reduce the time and labour cost of point-cloud annotation. However, unlike the existing weak-supervision methods, our \nickname{} does not require any self-supervised pretraining, hand-crafted constraints, or complicated post-processing steps, whilst obtaining close to fully-supervised accuracy using as few as 0.1\% training labels on multiple large-scale open datasets. Remarkably, for similar accuracy, we find that labelling costs (time) can be reduced up to 98\% according to our empirical evaluation in Appendix. Figure \ref{fig:teaser} shows the qualitative results of our method. Our key contributions are:
\begin{itemize}[leftmargin=*]
\setlength{\itemsep}{1pt}
\setlength{\parsep}{1pt}
\setlength{\parskip}{1pt}
    \item We propose a new weakly supervised method that leverages a point neighbourhood query to fully utilize the sparse training signals.
    \item We observe that existing fully-supervised methods degrade slowly until 1\% sparse point annotation, demonstrating that full, dense labelling is redundant and not necessary.
    \item We demonstrate a significant improvement over baselines in our benchmark, and surpass the state-of-the-art weak-supervision methods by large margins.
\end{itemize}

\section{Related Work}
\label{sec:related_work}
\subsection{Learning with Full Supervision}\label{sec:sup_nets}

\textbf{End-to-End Full Supervision.} With the availability of densely-annotated point cloud datasets \cite{hu2020towards,2D-3D-S,Semantic3D,behley2019semantickitti,NPM3D,varney2020dales,Toronto3D}, deep learning-based approaches have achieved unprecedented development in semantic segmentation in recent years.
The majority of existing approaches follow the standard end-to-end training strategy. They can be roughly divided into three categories according to the representation of 3D point clouds \cite{guo2019deep}: \textbf{1) Voxel-based methods.} They \cite{cheng20212,yan2020sparse,sparse,vvnet} usually voxelize the irregular 3D point clouds into regular cubes \cite{tchapmi2017segcloud,4dMinkpwski}, cylinders \cite{cylinder3d}, or spheres \cite{SPH3D}. \textbf{2) 2D Projection-based methods.} This pipeline projects the unstructured 3D points into 2D images through multi-view \cite{snapnet,kundu2020virtual}, bird-eye-view \cite{aksoy2019salsanet}, or spherical projections \cite{rangenet++,salsanext,wu2018squeezeseg,wu2019squeezesegv2,xu2020squeezesegv3}, and then uses the mature 2D architectures \cite{long2015fcn,he2016deep} for semantic learning. \textbf{3) Point-based methods.} These methods \cite{hu2019randla,qi2017pointnet,qi2017pointnet++,thomas2019kpconv,li2018pointcnn,wu2018pointconv,Point_transformer} directly operate on raw point clouds using shared MLPs. Hybrid representations, such as point-voxel representation \cite{e3d,Point_voxel_cnn,rethage2018fully}, 2D-3D representation \cite{feihu,jaritz2019multi}, are also studied.

\smallskip\noindent\textbf{Self-supervised Pretraining + Full Finetuning.}  Inspired by the success of self-supervised pre-training representation learning in 2D images \cite{simclr,he2020momentum}, several recent studies \cite{xie2020pointcontrast,liu2020p4contrast,wang2020pre,zhang2021self,thabet2020self,jigsaw,Jiang_2021_ICCV,Chen_2021_ICCV} apply contrastive techniques for 3D semantic segmentation. These methods usually pretrain the networks on additional 3D source datasets to learn initial per-point representations via self-supervised contrastive losses, after which the networks are carefully finetuned on the target datasets with full labels. This noticeably improves the overall accuracy.

Although these methods have achieved remarkable results on existing datasets, they rely on a large amount of labeled data for training, which is costly and prohibitive in real applications. By contrast, this paper aims to learn semantics from a small fraction of annotations, which is cheaper and more realistic in practice.

\subsection{Unsupervised Learning}

Saudar and Sievers \cite{jigsaw} learn the point semantics by recovering the correct voxel position of every 3D point after the point cloud is randomly shuffled. Sun et al. propose Canonical Capsules \cite{Canonical_Capsules} to decompose point clouds into object parts and elements via self-canonicalization and auto-encoding. Although they have obtained promising results, they are limited to simple objects and cannot process the complex large-scale point clouds.

\subsection{Learning with Weak Supervision}
\label{sec2.3}

\textbf{Limited Indirect Annotations.} Instead of having point-level semantic annotations, only sub-cloud level or seg-level labels are available. Wei et al. \cite{mprm} firstly train a classifier with sub-cloud labels, and then generate point-level pseudo labels using class activation mapping technique \cite{cam}. Tao et al. \cite{tao2020seggroup} present a grouping network to learn semantic and instance segmentation of 3D point clouds, with the seg-level labels generated by over-segmentation pre-processing. Ren et al. \cite{ren20213d} present a multi-task learning framework for both semantic segmentation and 3D object detection with scene-level tags.

\smallskip\noindent\textbf{Limited Point Annotations.} Given a small fraction of points with accurate semantic labels for training, Xu and Lee \cite{xu2020weakly} propose a weakly supervised point cloud segmentation method by approximating gradients and using handcrafted spatial and color smoothness constraints. 
Zhang et al. \cite{zhang2021perturbed} explicitly added a perturbed branch, and achieve weakly-supervised learning on 3D point clouds by enforcing predictive consistency. Shi et al. \cite{shi2021label_effcient} further investigate label-efficient learning by introducing a super-point-based active learning strategy. In addition, self-supervised pre-training methods \cite{self_few_shot,xie2020pointcontrast,hou2020exploring,zhang2021self,liu2020p4contrast,Zhang_2021_ICCV} are also flexible to fine-tune networks on limited annotations. Our \nickname{} is designed for limited point annotations which we believe has greater potential in practical applications. It does not require any pre-training, post-processing, or active labelling strategies, while achieving similar or even higher performance than the fully-supervised counterpart with only 0.1\% randomly annotated points for training.

\smallskip\noindent\textbf{Fair comparison with super-voxel based methods \cite{liu2021one,wu2022pointmatch} on ScanNet\footnote{In the previous version, we inadvertently provided some inaccurate descriptions. We apologize for the oversight and have corrected the information in this version.}.} 
In the interests of fair and reproducible comparison, we point out that a few published works claim state-of-the-art results yet rely on potentially flawed assumptions. Specifically,

\begin{itemize}
    \item \textbf{Inappropriate usage of the provided ScanNet segments as supervoxels}. 1T1C \cite{liu2021one} utilizes the segments provided by the ScanNet dataset as the super-voxel partition\footnote{\url{https://github.com/liuzhengzhe/One-Thing-One-Click/issues/13}}. However, this approach carries an implicit assumption: that the labels of the points in each supervoxel after over-segmentation are pure and consistent. While this assumption may hold true for ScanNet, as its labeling process is based on the provided segments, its validity for other datasets is questionable. In cases where the labels in each supervoxel are not pure and consistent, propagating the label to the entire voxel would introduce errors, which would in turn affect the network's training (as possibly evidenced by the decreased performance in the S3DIS dataset).
    \item \textbf{Misleading (ambiguous) labeling ratios.} 1T1C calculates its labeling ratio by using the number of clicks divided by the total number of raw points, resulting in a remarkably low labeling ratio (e.g., 0.02\%)\footnote{\url{https://github.com/liuzhengzhe/One-Thing-One-Click/issues/8}}. A fairer method, as used in prior art \cite{xu2020weakly,Zhang_2021_ICCV,zhang2021weakly}, is to use the total number of labeled points (\textit{i.e.,} to maintain consistency) divided by the total number of points. Considering that semantic annotations within each supervoxel are clean and consistent, 1 click per supervoxel is equivalent to labeling all points within a supervoxel. Consequently, the super-voxel semantic labels used by 1T1C are actually dense in ScanNet, while significantly larger than 0.02\%. Detailed analysis and discussions between us and the creators of ScanNet can be found at this \href{https://github.com/QingyongHu/SQN/blob/main/imgs/discussion_ScanNet.png}{Link}.
\end{itemize}
For these reasons, our method cannot directly compare with these methods on ScanNet. We kindly suggest that future research in this area addresses these issues, ensuring a more accurate and fair comparison between different approaches.

\section{Exploring Weak Supervision}
\label{sec:benchmark}

As weakly-supervised 3D semantic segmentation is still in its infancy, there is no consensus about what are the sensible formulations of weak training signals, and what approach should be used to sparsely annotate a dataset  such that a direct comparison is possible. We first explore this, then we investigate how existing fully supervised techniques perform under a weak labelling regime.

\textbf{Weak Annotation Strategy:} The fundamental objective of weakly-supervised segmentation is to obtain accurate estimations with as low as possible annotation cost, in terms of labeller time. However, it is non-trivial to compare the cost of different annotation methods in practice. Existing annotation options include 1) randomly annotating sparse point labels \cite{xu2020weakly,zhang2021perturbed,zhang2021weakly}, 2) actively annotating sparse point labels \cite{hou2020exploring,shi2021label_effcient} or region-wise labels \cite{wu2021redal}, 3) annotating seg-level labels or superpoint labels \cite{tao2020seggroup,liu2021one,cheng2021sspc} and 4) annotating sub-cloud labels \cite{mprm}.  All methods have merits. For the purpose of fair reproducibility, we opt for the random point annotation strategy, considering the practical simplicity of building such an annotation tool.

\begin{figure}[t]
	\begin{center}
		\includegraphics[width=1.0\textwidth]{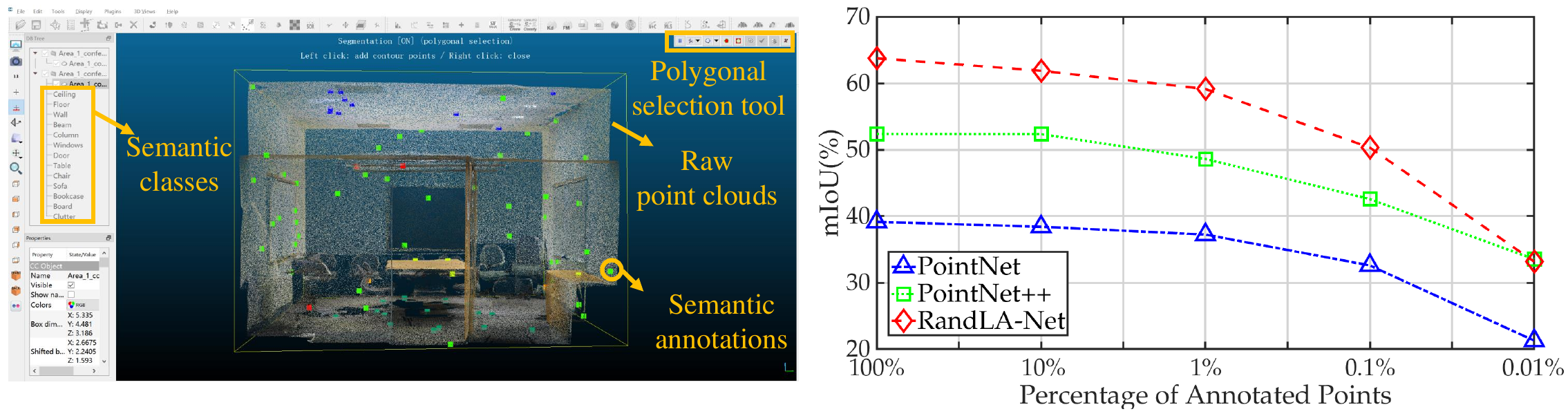}
	\end{center}
    \caption{Left: Illustration of the sparse annotation tool. Right: Degradation of three baselines in the \textit{Area-5} of S3DIS \cite{2D-3D-S} when decreasing proportions of points that are randomly annotated. (Logarithmic scale used in horizontal axis).}
	\label{fig: critical_point}
\vspace{-.2cm}
\end{figure}

\textbf{Annotation Tool:} To verify the feasibility of random sparse annotations in practice, we develop a user-friendly labelling pipeline based on the off-the-shelf CloudCompare\footnote{https://www.cloudcompare.org/} software. Specifically, we first import raw 3D point clouds to the software and randomly downsample them to 10\%/1\%/0.1\% of the total points for sparse annotation. Considering the sparsity of the remaining points, we explicitly enlarge the size of selected points and take the original full point clouds as a reference. As illustrated in left part of Figure \ref{fig: critical_point}, we then use the standard labelling mode such as polygonal edition for point-wise annotating. (Details and video recordings of our annotation pipeline are supplied in the appendix).

\textbf{Annotation Cost:} 
With the developed annotation tool, it takes less than 2 minutes to annotate 0.1\% of points of a standard room in the S3DIS dataset. For comparison, it requires more than 20 minutes to fully annotate all points for the same room. Note that, the sparse annotation scheme is particularly suitable for large-scale 3D point clouds with billions of points. As detailed in the appendix, it only takes about 18 hours to annotate 0.1\% of the urban-scale SensatUrban dataset \cite{hu2020towards}, while annotating all points requires more than 600 person-hours. 

\textbf{Experimental Settings:} We choose the well-known S3DIS dataset \cite{2D-3D-S} as the testbed. The Areas $\{1/2/3/4/6\}$ are selected as the training point clouds, the Area 5 is fully annotated for testing only. With the random sparse annotation strategy, we set up the following four groups of weak signals for training. Specifically, we only annotate the randomly selected 10\%/1\%/0.1\%/0.01\% of the 3D points in each room in all training areas.

\textbf{Using Fully-supervised Methods as Baselines.} We select the seminal works PointNet/PointNet++ \cite{qi2017pointnet,qi2017pointnet++} and the recent large-scale-point-cloud friendly RandLA-Net \cite{hu2019randla} as baselines. These methods are end-to-end trained on the four groups of weakly annotated data without using any additional modules. During training, only the labeled points are used to compute the loss for back-propagation. In total, 12 models (3 models/group $\times$ 4 groups) are trained for evaluation on the full Area 5. Detailed results can be found in Appendix.

\textbf{Results and Findings.}
Figure \ref{fig: critical_point} shows the mIoU scores of all models for segmenting the total 13 classes. The results under full supervision (100\% annotations for all training data) are included for comparison. It can be seen that:
\begin{itemize}[leftmargin=*]
\setlength{\itemsep}{0pt}
\setlength{\parsep}{0pt}
\setlength{\parskip}{0pt}
    \item The performance of all baselines only decreases marginally (less than 4\%) even though the proportion of point annotations drops significantly from 100\% to 1\%.
    This clearly shows that the dense annotations are actually unnecessary to obtain a comparable and favorable segmentation accuracy under the simple random annotation strategy. 
    \item The performance of all baselines drops significantly once the annotated points are lower than 0.1\%. This critical point indicates that keeping a certain amount of training signals is also essential for weak supervision.
\end{itemize}

Above all, we may conclude that for segmenting large-scale point clouds which are usually dominated by major classes and have numerous repeatable local patterns, it is desirable to develop weakly-supervised methods which have an excellent trade-off between annotation costs and estimation accuracy. With this motivation, we propose \nickname{} which achieves close to fully-supervised accuracy using only 0.1\% labels for training.

\section{\nickname{}}
\label{sec:method}

\subsection{Overview}
Given point clouds with sparse annotations,
the fundamental challenge for weakly-supervised learning is how to fully utilize the sparse yet valuable training signals to update the network parameters, such that more geometrically meaningful local patterns can be learned. To resolve this, we design a simple \nickname{} which consists of two major components: 1) a point local feature extractor to learn diverse visual patterns; 2) a flexible point feature query network to collect as many as possible relevant semantic features for weakly-supervised training. 
As shown in Figure \ref{fig: pipeline}, our two sub-networks are illustrated by the stacked blocks. 

\subsection{Point Local Feature Extractor}
This component aims to extract local features for all points. As discussed in Section \ref{sec:sup_nets}, there are many excellent backbone networks that are able to extract per-point features. In general, these networks stack multiple encoding layers together with downsampling operations to extract hierarchical local features. In this paper, we use the encoder of RandLA-Net \cite{hu2019randla} as our feature extractor thanks to its efficiency on large-scale point clouds. Note that \nickname{} is not restricted to any particular backbone network \textit{e.g.} as we demonstrate in the Appendix with MinkowskiNet~\cite{4dMinkpwski}. 

\begin{figure}[t]
	\begin{center}
		\includegraphics[width=0.95\textwidth]{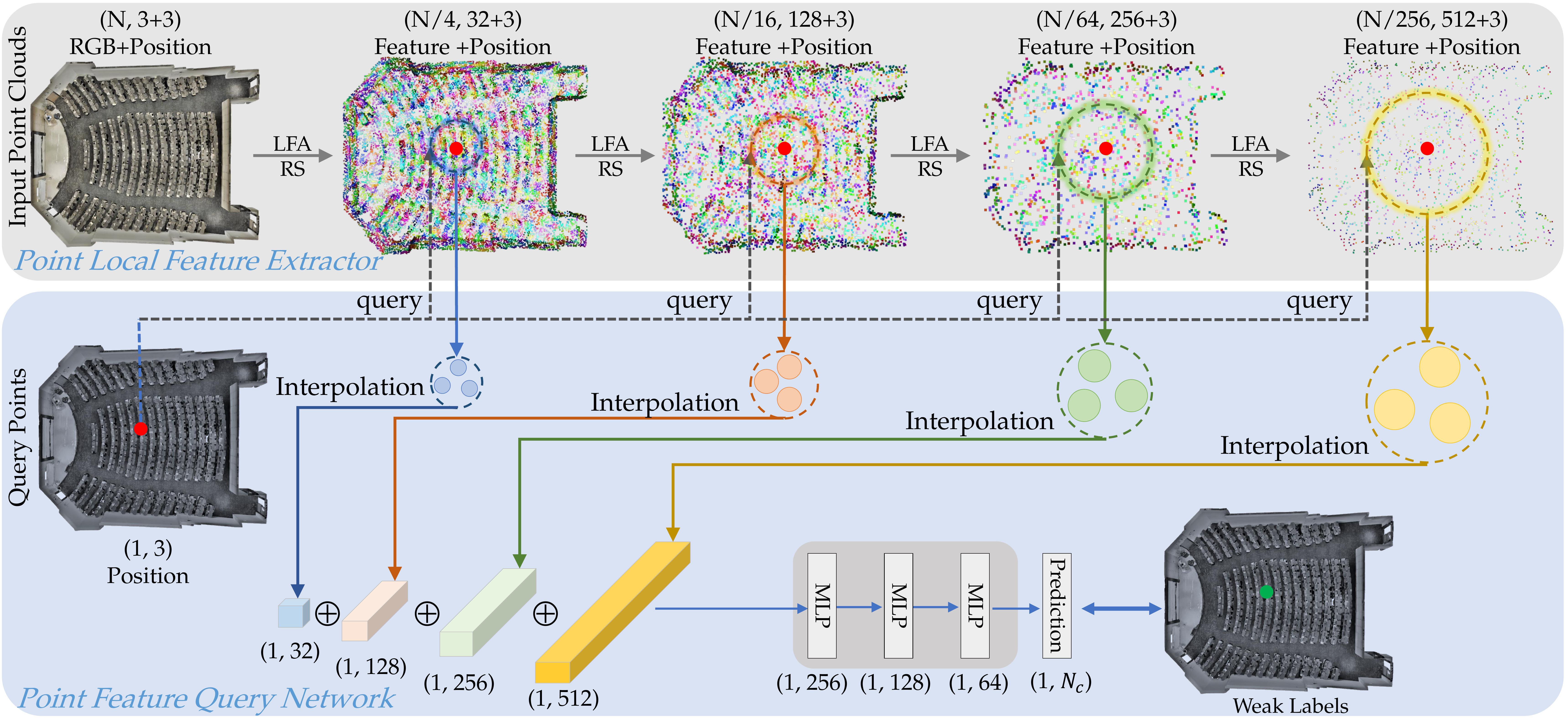}
	\end{center}
    \caption{The pipeline of our \nickname{} at the training stage with weak supervision. We only show one query point for simplicity. }
	\label{fig: pipeline}
\end{figure}

As shown in the top block of Figure \ref{fig: pipeline}, the encoder includes four layers of Local Feature Aggregation (LFA) followed by a Random Sampling (RS) operation. Details refer to RandLA-Net \cite{hu2019randla}. Given an input point cloud $\mathcal{P}$ with $N$ points, four levels of hierarchical point features are extracted after each encoding layer, \textit{i}.\textit{e}., 1) $\frac{N}{4} \times 32$, 2)$\frac{N}{16}\times 128$, 3) $\frac{N}{64}\times 256$, and 4) $\frac{N}{256}\times 512$. To facilitate the subsequent query network, the corresponding point location $xyz$ are always preserved for each hierarchical feature vector.

\subsection{Point Feature Query Network}
Given the extracted point features, this query network is designed to collect as many relevant features, to be trained using the available sparse signals. In particular, as shown in the bottom block of Figure \ref{fig: pipeline}, it takes a specific 3D query point as input and then acquires a set of learned point features relevant to that point. Fundamentally, this is assumed that the query point shares similar semantic information with the collected point features, such that the training signals from the query points can be shared and back-propagated for the relevant points. The network consists of: 1) Searching Spatial Neighbouring Point Features, 2) Interpolating Query Point Features, 3) Inferring Query Point Semantics.

\begin{figure}[thb]
	\begin{center}
		\includegraphics[width=0.9\textwidth]{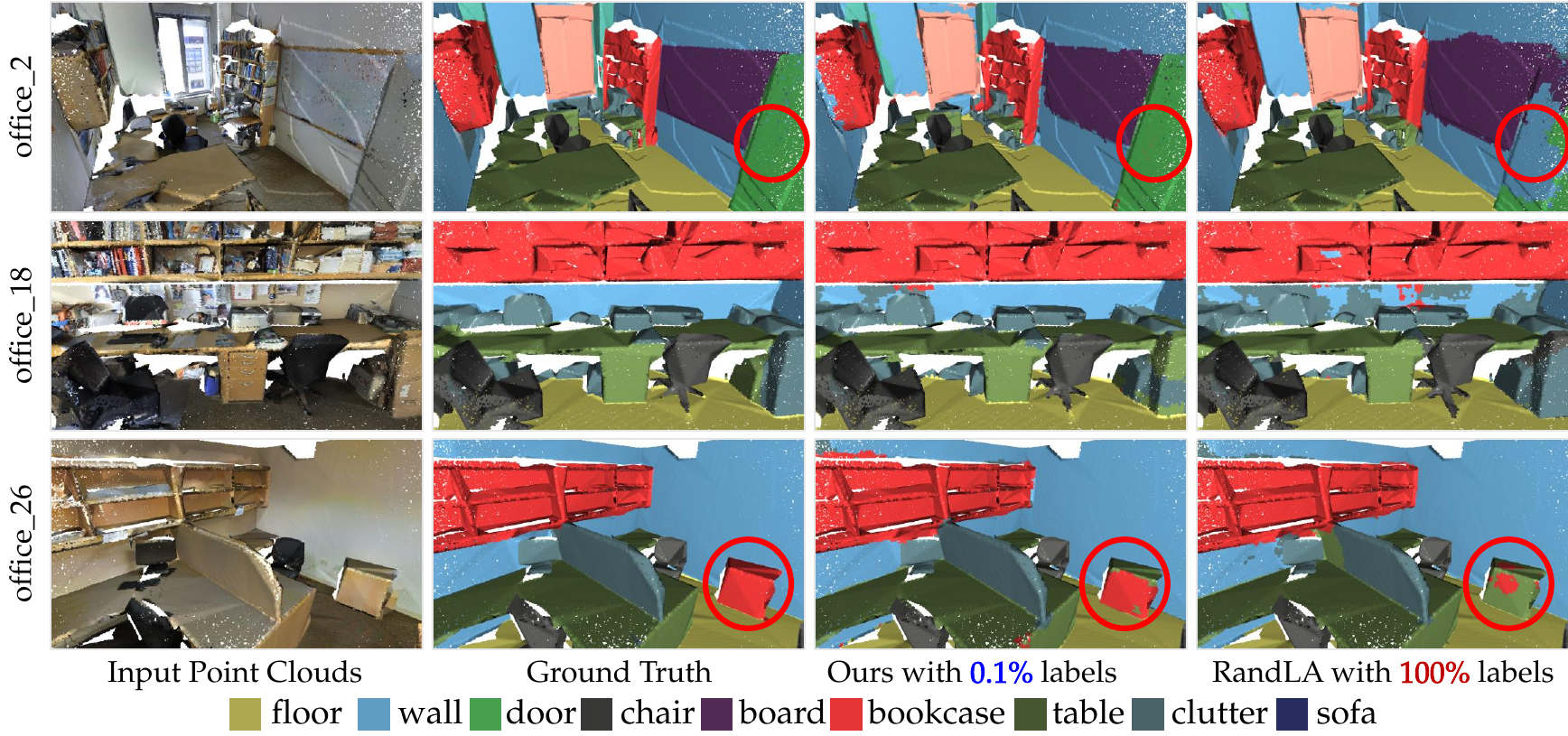}
	\end{center}
	\vspace{-0.7cm}
    \caption{Qualitative results achieved by our \nickname{} and the fully-supervised RandLA-Net \cite{hu2019randla} on the \textit{Area-5} of the S3DIS dataset.}
	\label{fig: S3DIS_qualitative}
\end{figure}
	\vspace{-0.5cm}

\textbf{Searching Spatial Neighbouring Point Features.}
Given a 3D query point $p$ with its location $xyz$, this module is to simply search the nearest $K$ points in each of the previous 4-level encoded features, according to the point-wise Euclidean distance. For example, as to the first level of extracted point features, the most relevant $K$ points are selected, acquiring the raw features \{$\boldsymbol{F}^1_p, \dots \boldsymbol{F}^K_p$\}.

\textbf{Interpolating Query Point Features.}
For each level of features, the queried $K$ vectors are compressed into a compact representation for the query point $p$. For simplicity, we apply the trilinear interpolation method to compute a feature vector for $p$, according to the Euclidean distance between $p$ and each of $K$ points.  Eventually, four hierarchical feature vectors are concatenated together, representing all relevant point features from the entire 3D point cloud.

\textbf{Inferring Query Point Semantics.}
After obtaining the unique and representative feature vector for the query point $p$, we feed it into a series of MLPs, directly inferring the point semantic category. 

Overall, given a sparse number of annotated points, we query their neighbouring point features in parallel for training. This allows the valuable training signals to be back-propagated to a much wider spatial context. During testing, all 3D points are fed into the two sub-networks for semantic estimation. In fact, our simple query mechanism allows the network to infer the point semantic category from a significantly larger receptive field.

\vspace{-0.5cm}
\subsection{Implementation Details}
\vspace{-0.2cm}

The hyperparameter $K$ is empirically set to 3 for semantic query in our framework and \textbf{kept consistent for all experiments}. Our \nickname{} follows the dataset preprocessing used in RandLA-Net \cite{hu2019randla}, and is trained end-to-end with 0.1\% randomly annotated points. All experiments are conducted on a PC with an Intel Core™ i9-10900X CPU and an NVIDIA RTX Titan GPU. Note that, the proposed SQN framework allows flexible use of different backbone networks such as voxel-based MinkowskiNet \cite{4dMinkpwski}, please refer to the appendix for more details.

\begin{table}[t]
\centering
\resizebox{\textwidth}{!}{%
\begin{tabular}{c|rcccccccccccccc}
\Xhline{2.0\arrayrulewidth}
\multicolumn{1}{l|}{} & Methods & \textbf{mIoU(\%)} & ceiling & floor & wall & beam & column & window & door & table & chair & sofa & bookcase & board & clutter \\ 
\Xhline{2.0\arrayrulewidth}
\multirow{7}{*}{\begin{tabular}[c]{@{}c@{}}Full\\ supervision\end{tabular}} 
& PointNet \cite{qi2017pointnet} & 41.1 & 88.8 & 97.3 & 69.8 & 0.1 & 3.9 & 46.3 & 10.8 & 58.9 & 52.6 & 5.9 & 40.3 & 26.4 & 33.2 \\
 & PointCNN \cite{li2018pointcnn} & 57.3 & 92.3 & 98.2 & 79.4 & 0.0 & 17.6 & 22.8 & 62.1 & 74.4 & 80.6 & 31.7 & 66.7 & 62.1 & 56.7 \\
 & SPGraph \cite{landrieu2018large} & 58.0 & 89.4 & 96.9 & 78.1 & 0.0 & {\ul 42.8} & 48.9 & 61.6 & {\ul 84.7} & 75.4 & 69.8 & 52.6 & 2.1 & 52.2 \\
 & SPH3D \cite{SPH3D} & 59.5 & {\ul 93.3} & 97.1 & 81.1 & 0.0 & 33.2 & 45.8 & 43.8 & 79.7 & 86.9 & 33.2 & 71.5 & 54.1 & 53.7 \\
 & PointWeb \cite{pointweb} & 60.3 & 92.0 & {\ul 98.5} & 79.4 & 0.0 & 21.1 & 59.7 & 34.8 & 76.3 & 88.3 & 46.9 & 69.3 & 64.9 & 52.5 \\
  & RandLA-Net \cite{hu2019randla} & 63.0 & 92.4 & 96.7 & 80.6 & 0.0 & 18.3 & {\ul 61.3} & 43.3 & 77.2 & 85.2 & {\ul 71.5} & 71.0 & {\ul 69.2} & 52.3 \\
 & KPConv rigid \cite{thomas2019kpconv} & {\ul 65.4} & 92.6 & 97.3 & {\ul 81.4} & 0.0 & 16.5 & 54.5 & {\ul 69.5} & 80.2 & {\ul 90.1} & 66.4 & {\ul 74.6} & 63.7 & {\ul 58.1} \\
\Xhline{1.0\arrayrulewidth}
\multirow{2}{*}{\begin{tabular}[c]{@{}c@{}}Limited \\ superpoint labels$^\dagger$\end{tabular}} 
& 1T1C (0.02\%) \cite{liu2021one} & 50.1 & - & - & - & - & - & - & - & - & - & - & - & - & - \\
 & SSPC-Net (0.01\%) \cite{cheng2021sspc} & 51.5 & - & - & - & - & - & - & - & - & - & - & - & - & - \\
\Xhline{1.0\arrayrulewidth}
\multirow{10}{*}{\begin{tabular}[c]{@{}c@{}}Limited \\ point-wise\\ labels\end{tabular}} 
& $\Pi$ Model (10\%) \cite{laine2016temporal} & 46.3 & 91.8 & 97.1 & 73.8 & 0.0 & 5.1 & 42.0 & 19.6 & 67.2 & 66.7 & 47.9 & 19.1 & 30.6 & 41.3 \\
 & MT (10\%) \cite{MT} & 47.9 & 92.2 & 96.8 & 74.1 & 0.0 & 10.4 & 46.2 & 17.7 & 70.7 & 67.0 & 50.2 & 24.4 & 30.7 & 42.2 \\
 & Xu (10\%) \cite{xu2020weakly} & 48.0 & 90.9 & 97.3 & 74.8 & 0.0 & 8.4 & 49.3 & 27.3 & 71.7 & 69.0 & 53.2 & 16.5 & 23.3 & 42.8 \\ \cdashline{2-16} 
 & Zhang et al. (1\%) \cite{zhang2021weakly} & 61.8 & 91.5 & 96.9 & 80.6 & 0.0 & 18.2 & 58.1 & 47.2 & 75.8 & 85.7 & 65.2 & 68.9 & 65.0 & 50.2 \\
 & PSD (1\%) \cite{zhang2021perturbed} & 63.5 & 92.3 & 97.7 & 80.7 & 0.0 & 27.8 & 56.2 & 62.5 & 78.7 & 84.1 & 63.1 & 70.4 & 58.9 & 53.2 \\ \cdashline{2-16} 
 & $\Pi$ Model (0.2\%) \cite{laine2016temporal} & 44.3 & 89.1 & 97.0 & 71.5 & 0.0 & 3.6 & 43.2 & 27.4 & 63.1 & 62.1 & 43.7 & 14.7 & 24.0 & 36.7 \\
 & MT (0.2\%) \cite{MT}  & 44.4 & 88.9 & 96.8 & 70.1 & 0.1 & 3.0 & 44.3 & 28.8 & 63.7 & 63.6 & 47.7 & 15.5 & 23.0 & 35.8 \\
 & Xu (0.2\%) \cite{xu2020weakly} & 44.5 & 90.1 & 97.1 & 71.9 & 0.0 & 1.9 & 47.2 & 29.3 & 64.0 & 62.9 & 42.2 & 15.9 & 18.9 & 37.5 \\ \cdashline{2-16} 
 & RandLA-Net (0.1\%) & 52.9 & 89.9 & 95.9 & 75.3 & 0.0 & 7.5 & 52.4 & 26.5 & 62.2 & 74.5 & 49.1 & 60.2 & 49.3 & 45.1 \\
 & \textbf{Ours (0.1\%)} & \textbf{61.4} & \textbf{91.7} & 95.6 & \textbf{78.7} & 0.0 & \textbf{24.2} & \textbf{55.9} & \textbf{63.1} & \textbf{70.5} & \textbf{83.1} & \textbf{60.7} & \textbf{67.8} & \textbf{56.1} & \textbf{50.6} \\ 
\Xhline{2.0\arrayrulewidth}
\end{tabular}%
}
\caption{Quantitative results of different methods on the \textit{Area-5} of S3DIS dataset. Mean IoU (mIoU, \%), and per-class IoU (\%) scores are reported. Bold represents the best result in weakly labelled settings and underlined represents the best under fully labelled settings. $^\dagger$As mentioned in Sec. \ref{sec2.3}, misleading labeling ratio is reported, and hence a direct comparison is not possible.}
\label{tab:S3DIS}
\end{table}

\vspace{-0.3cm}

\section{Experiments}
\label{Sec:Experiments}

\subsection{Comparison with SOTA Approaches}
\label{subsec:Comparison_with_SOTA}

We first evaluate the performance of our \nickname{} on three commonly-used benchmarks including S3DIS \cite{2D-3D-S}, ScanNet \cite{scannet} and Semantic3D \cite{Semantic3D}. Following \cite{hu2019randla}, we use the Overall Accuracy (OA) and mean Intersection-over-Union (mIoU) as the main evaluation metrics.

\smallskip\noindent\textbf{Evaluation on S3DIS.}\enskip Following \cite{xu2020weakly}, we report the results on Area-5 in Table \ref{tab:S3DIS}. Note that, our SQN is compared with three groups of approaches: 1)  Fully-supervised methods including SPGraph \cite{landrieu2018large}, KPConv \cite{thomas2019kpconv} and RandLA-Net with 100\% training labels; 2) Weakly supervised approaches that learn from limited superpoint annotations including 1T1C \cite{liu2021one} and SSPC-Net \cite{cheng2021sspc}; 3) Weakly-supervised methods \cite{xu2020weakly,MT,laine2016temporal} that learning from limited annotations. We also list the proportion of annotations used for training.

Considering different backbones and different labelling ratios are used by existing methods, we focus on the comparison of our SQN and the baseline RandLA-Net, which under the same weakly-supervised settings. It can be seen that our \nickname{} outperforms RandLA-Net by nearly 9\% under the same 0.1\% random sparse annotations. In particular, our \nickname{} is also comparable to the fully-supervised RandLA-Net \cite{hu2019randla}. Figure \ref{fig: S3DIS_qualitative} shows qualitative comparisons of RandLA-Net and our \nickname{}.

\begin{table}[t]
\parbox{.4\textwidth}{
\centering
\resizebox{0.45\textwidth}{!}{%
\begin{tabular}{c|rc}
\Xhline{2\arrayrulewidth}
Settings & Methods & \textbf{mIoU(\%)} \\
\hline
\multirow{7}{*}{\begin{tabular}[c]{@{}c@{}}Full\\ supervision\end{tabular}}
& PointNet++ \cite{qi2017pointnet++} & 33.9 \\
 & SPLATNet \cite{su2018splatnet} & 39.3 \\
 & TangentConv \cite{tangentconv} & 43.8 \\
 & PointCNN \cite{li2018pointcnn} & 45.8 \\
 & PointConv \cite{wu2018pointconv} & 55.6 \\
 & SPH3D-GCN \cite{SPH3D} & 61.0 \\
 & KPConv \cite{thomas2019kpconv} & 68.4 \\
 & RandLA-Net \cite{hu2019randla} & 64.5 \\
 \hline
\multirow{4}{*}{\begin{tabular}[c]{@{}c@{}}Weak\\ supervision\end{tabular}} 
 & MPRM* \cite{mprm} & 41.1 \\
 & Zhang \textit{et al.} (1\%) \cite{zhang2021weakly} & 51.1 \\
  & PSD (1\%) \cite{zhang2021perturbed} & 54.7 \\
 & \textbf{\nicknamenew{} (0.1\%)} &  \textbf{56.9}\\
\Xhline{2\arrayrulewidth}
\end{tabular}}
\caption{Quantitative results on ScanNet (online test set). *MPRM \cite{mprm} takes sub-cloud labels as supervision signal.}
\label{tab:scannet}
}
\hfill
\parbox{.5\textwidth}{
\centering
\resizebox{0.5\textwidth}{!}{%
\begin{tabular}{c|rcccc}
\Xhline{2\arrayrulewidth}
 &  & \multicolumn{2}{c}{\textit{Semantic8}} & \multicolumn{2}{c}{\textit{Reduced8}} \\ 
 & Methods & OA(\%) & \textbf{mIoU(\%)} & OA(\%) & \textbf{mIoU(\%)} \\
\hline
\multirow{7}{*}{\begin{tabular}[c]{@{}c@{}}Full\\ sup.\end{tabular}} 
 & SnapNet \cite{snapnet} & 91.0 & 67.4 & 88.6 & 59.1 \\
 & PointNet++ \cite{qi2017pointnet++} & 85.7 & 63.1 & - & - \\ 
 & ShellNet \cite{zhang2019shellnet} & - & - & 93.2 & 69.3 \\
 & GACNet \cite{GACNet} & - & - & 91.9 & 70.8 \\
 & RGNet \cite{RGNet} & 90.6 & 72.0 & 94.5 & 74.7 \\
 & SPG \cite{landrieu2018large} & 92.9 & 76.2 & 94.0 & 73.2 \\
 & KPConv \cite{thomas2019kpconv} & - & - & 92.9 & 74.6 \\
 & ConvPoint \cite{boulch2020lightconvpoint} & 93.4 & 76.5 & - & - \\
 & WreathProdNet \cite{wang2020equivariant} & 94.6 & \uline{77.1} & - & - \\
 & RandLA-Net \cite{hu2019randla} & \uline{95.0} & 75.8 & \uline{94.8} & \uline{77.4} \\
\hline
\multirow{4}{*}{\begin{tabular}[c]{@{}c@{}}Weak\\ sup.\end{tabular}} 
 & Zhang \textit{et al.} (1\%) \cite{zhang2021weakly} & - & - & - & 72.6 \\
  & PSD (1\%) \cite{zhang2021perturbed} & - & - & - & 75.8 \\
 & \textbf{\nicknamenew{} (0.1\%)} & \textbf{94.8} & \textbf{72.3} & \textbf{93.7} & \textbf{74.7} \\
 & \textbf{\nicknamenew{} (0.01\%)} & 91.9  & 58.8 & 90.3 & 65.6 \\
 \Xhline{2.0\arrayrulewidth}
\end{tabular}}%
\caption{Quantitative results on Semantic3D \cite{Semantic3D}. The scores are obtained from the recent publications.}
\label{tab:Semantic3D}
}
\end{table}

\smallskip\noindent\textbf{Evaluation on ScanNet.}\enskip We report the quantitative results achieved by different approaches on the hidden test set in Table \ref{tab:scannet} . It can be seen that our \nickname{} achieves higher mIoU scores with only 0.1\% training labels, compared with MPRM \cite{mprm} which is trained with sub-cloud labels, and Zhang et al. \cite{zhang2021weakly} and PSD \cite{zhang2021perturbed} trained with 1\% annotations. Considering that the actual training settings in the ScanNet Data-Efficient benchmark cannot be verified, hence we do not provide the comparison in this benchmark.

\smallskip\noindent\textbf{Evaluation on Semantic3D.} \enskip Table \ref{tab:Semantic3D} compares our SQN with a number of fully-supervised methods. It can be seen that our \nickname{} trained with 0.1\% labels achieves competitive performance with fully-supervised baselines on both \textit{Semantic8} and \textit{Reduced8} subsets. This clearly demonstrates the effectiveness of our semantic query framework, which takes full advantage of the limited annotations. Additionally, we also train our \nickname{} with only 0.01\% randomly annotated points, considering the extremely large amount of 3D points scanned. We can see that our \nickname{} trained with 0.01\% labels also achieves satisfactory accuracy, though there is space to be improved in the future.

\subsection{Evaluation on Large-Scale 3D Benchmarks}
\label{subsec:Large-Scale_3D_Benchmarks}

To validate the versatility of our \nickname{}, we further evaluate our \nickname{} on four point cloud datasets with different density and quality, including SensatUrban \cite{hu2020towards}, Toronto3D \cite{Toronto3D}, DALES \cite{varney2020dales}, and SemanticKITTI \cite{behley2019semantickitti}. Note that, all existing weakly supervised approaches are only evaluated on the dataset with dense point clouds, and there are no results reported on these datasets. Therefore, we only compare our approach with existing fully-supervised methods in this section.

\begin{table}[t]
\centering
\resizebox{\textwidth}{!}{%
\begin{tabular}{c|rcc|ccc|cc|c}
\Xhline{2\arrayrulewidth}
\multirow{2}{*}{Settings} & \multirow{2}{*}{Methods} & \multicolumn{2}{c}{DALES \cite{varney2020dales}} & \multicolumn{3}{c}{SensatUrban \cite{hu2020towards}} & \multicolumn{2}{c}{Toronto3D \cite{Toronto3D}} & SemanticKITTI \cite{behley2019semantickitti} \\ \cline{3-10} 
 &  & OA(\%) & \textbf{mIoU(\%)} & OA(\%) & mAcc (\%) & \textbf{mIoU(\%)} & OA(\%) & \textbf{mIoU(\%)} & \textbf{mIoU(\%)} \\ 
 \Xhline{2\arrayrulewidth}
\multirow{11}{*}{Full supervision} 
& PointNet \cite{qi2017pointnet} & - & - & 80.8 & 30.3 & 23.7 & - & - & 14.6 \\
 & PointNet++ \cite{qi2017pointnet++} & 95.7 & 68.3 & 84.3 & 40.0 & 32.9 & 84.9 & 41.8 & 20.1 \\
 & PointCNN \cite{li2018pointcnn} & 97.2 & 58.4 & - & - & - & - & - & - \\
 & TangentConv \cite{tangentconv} & - & - & 77.0 & 43.7 & 33.3 & - & - & 40.9 \\
 & ShellNet \cite{zhang2019shellnet} & 96.4 & 57.4 & - & - & - & - & - & - \\
 & DGCNN \cite{dgcnn} & - & - & - & - & - & 94.2 & 61.8 & - \\
 & SPG \cite{landrieu2018large} & 95.5 & 60.6 & 85.3 & 44.4 & 37.3 & - & - & 17.4 \\
 & SparseConv \cite{sparse} & - & - & 88.7 & 63.3 & 42.7 & - & - & - \\
 & KPConv \cite{thomas2019kpconv} & \uline{97.8} & \uline{81.1} & \uline{93.2} & 63.8 & \uline{57.6} & \uline{95.4} & 69.1 & \uline{58.1} \\
 & ConvPoint \cite{conv_pts} & 97.2 & 67.4 & - & - & - & - & - & - \\
 & RandLA-Net \cite{hu2019randla} & 97.1 & 80.0 & 89.8 & \uline{69.6} & 52.7 & 92.9 & \uline{77.7} & 53.9 \\ \hline
\multirow{2}{*}{\begin{tabular}[c]{@{}c@{}}Weak\\ supervision\end{tabular}} 
& \textbf{Ours (0.1\%)} & 97.0 & 72.0 & \textbf{91.0} & \textbf{70.9} & \textbf{54.0} & 96.7 & \textbf{77.7} & 50.8 \\
 & \textbf{Ours (0.01\%)} & 95.9 & 60.4 & 85.6 & 49.4 & 37.2 & 94.2 & 68.2 & 39.1 \\ 
 \Xhline{2\arrayrulewidth}
\end{tabular}%
}
\caption{Quantitative results of different approaches on the DALES \cite{varney2020dales}, SensatUrban \cite{hu2020towards}, Toronto3D \cite{Toronto3D} and SemanticKITTI \cite{behley2019semantickitti}. }
\label{tab:all_dataset}
\end{table}

As shown in Table \ref{tab:all_dataset}, the performance of our \nickname{} is on par with the fully-supervised counterpart RandLA-Net on several datasets, whilst the model is only supplied with 0.1\% labels for training. In particular, our SQN trained with 0.1\% labels even outperforms the fully supervised RandLA-Net on the SensatUrban dataset. This shows the great potential of our method, especially for extremely large-scale point clouds with billions of points, where the manual annotation is unrealistic and impractical. The detailed results can be found in Appendix.

\subsection{Ablation Study}
\label{Sec:ablation}

To evaluate the effectiveness of each module in our framework, we conduct the following ablation studies. All ablated networks are trained on  Areas$\{1/2/3/4/6\}$ with 0.1\% labels, and tested on the \textit{Area-5} of the S3DIS dataset.

\smallskip\noindent\textbf{(1) Variants of Semantic Queries.} The hierarchical point feature query mechanism 
is the major component of our \nickname{}. To evaluate this component, we perform semantic query at different encoding layers. In particular, we train four additional models, each of which has a different combination of queried neighbouring point features. 

From Table \ref{tab:query_layers} we can see that the segmentation performance drops significantly if we only collect the relevant point features at a single layer (\egours the first or the last layer), whilst querying at the last layer can achieve much better results than in the first layer. This is because the points in the last encoding layer are quite sparse but representative, aggregating a large number of neighboring points. Additionally, querying at different encoding layers and combining them is likely to achieve better segmentation results, mainly because it integrates different spatial levels of semantic content and considers more neighboring points.
\vspace{-0.5cm}

\begin{table}[htb]
\centering
\resizebox{0.5\textwidth}{!}{%
\begin{tabular}{c|cccc|cc}
\Xhline{2.0\arrayrulewidth}
Model & 1st & 2nd & 3rd & 4st & OA(\%) & mIoU(\%) \\ 
\hline
A & $\checkmark$  &  &  &  &  48.66 & 22.89 \\
B &  &  &  & $\checkmark$ & 75.54  & 46.02 \\
C & $\checkmark$ & $\checkmark$ &  &   & 70.76 & 38.18 \\
D & $\checkmark$ & $\checkmark$ & $\checkmark$ &  & 82.37 & 54.21 \\
E & $\checkmark$ & $\checkmark$ & $\checkmark$ & $\checkmark$ &  \textbf{86.15} & \textbf{61.41} \\
\Xhline{2.0\arrayrulewidth}
\end{tabular}%
}
\caption{Ablations of different levels of semantic query.}
\label{tab:query_layers}
\end{table}
\vspace{-0.5cm}

\begin{minipage}{\textwidth}
  \begin{minipage}[b]{0.45\textwidth}
    \centering
	\includegraphics[width=1.0\textwidth]{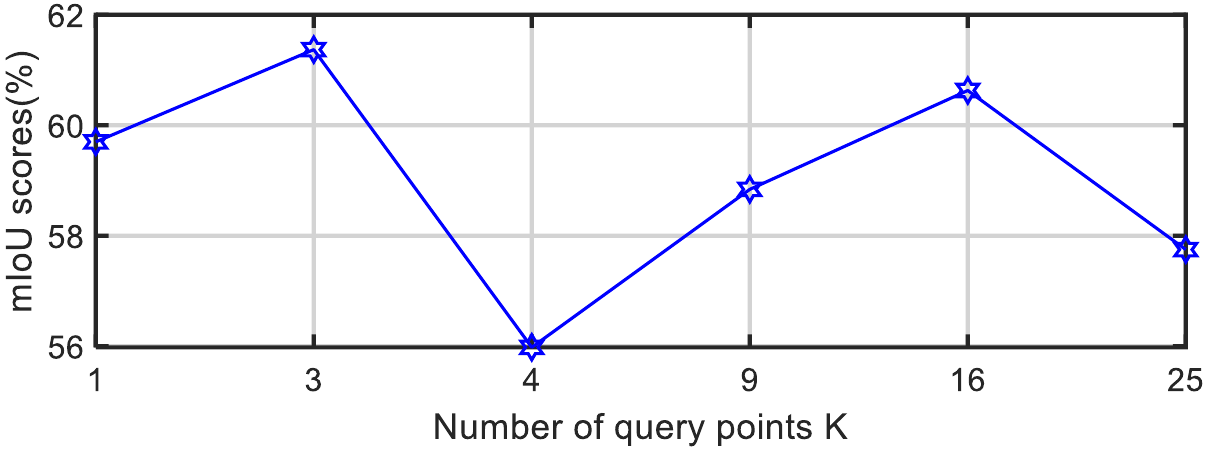}
    \captionof{figure}{The results of our \nickname{} with different number of query points on the \textit{Area-5} of the S3DIS dataset.}
    \label{fig: varying_query_points}
  \end{minipage}
  \hfill
  \begin{minipage}[b]{0.48\textwidth}
    \centering
    \begin{tabular}{ccc}
        \Xhline{2.0\arrayrulewidth}
         & OA(\%) & mIoU(\%) \\
        \Xhline{2.0\arrayrulewidth}
        Trial1 & 86.15 & 61.41 \\
        Trial2 & 85.63 & 59.24 \\
        Trial3 & 86.39 & 60.93 \\
        Trial4 & 86.32 & 59.40 \\
        Trial5 & \textbf{86.40} &
        \textbf{61.56} \\
        Mean & 86.25 & 60.42 \\
        STD & 0.32 & 0.93 \\
        \Xhline{2.0\arrayrulewidth}
        \end{tabular}%
      \captionof{table}{Sensitivity analysis of the proposed \nickname{} on S3DIS dataset (\textit{Area 5}) over 5 runs. }
      \label{tab:sensitivity}
    \end{minipage}
  \end{minipage}
\vspace{0.1cm}

\smallskip\noindent\textbf{(2) Varying Number of Queried Neighbours.} Intuitively, querying a larger neighborhood is more likely to achieve better results. However, an overly large neighborhood may include points with very different semantics, diminishing overall performance. To investigate the impact of the number of neighboring points used in our semantic query, we conduct experiments by varying the number of neighboring points from 1 to 25. As shown in Figure \ref{fig: varying_query_points}, the overall performance with differing numbers of neighboring points does not change significantly, showing that our simple query mechanism is robust to the size of the neighboring patch. Instead, the mixture of different feature levels plays a more important role as demonstrated in Table \ref{tab:query_layers}.

\smallskip\noindent\textbf{(3) Varying annotated points.} To verify the sensitivity of our \nickname{} to different randomly annotated points, we train our models five times with exactly the same architectures, \ieours the only change is that different subsets of randomly selected 0.1\% of points are labeled. The experimental results are reported in Table \ref{tab:sensitivity}. It can be seen that there are slight, but not significant, differences between different runs. This indicates that the proposed SQN is robust to the choice of randomly annotated points. We also notice that the major performance change lies in minor categories such as \textit{door}, \textit{sofa}, and \textit{board}, showing that the underrepresented classes are more sensitive to weak annotation. Please refer to appendix for details.

\smallskip\noindent\textbf{(4) Varying proportion of annotated points.} We further examine the performance of \nickname{} with differing amounts of annotated points. As shown in Table \ref{tab:annotation}, the proposed \nickname{} can achieve satisfactory segmentation performance when there are only 0.1\% labels available, but the performance drops significantly when there are only 0.01\% labeled points available, primarily because the supervision signal is too sparse and limited in this case. It is also interesting to see that our framework achieves slightly better mIoU performance when using 10\% labels compared with full supervision. In particular, the performance on minority categories such as \textit{column/window/door} has improved by 2\%-5\%. This implies that: 1) In a sense, the supervision signal is sufficient in this case; 2) Another way to address the critical issue of imbalanced class distribution may be to use a portion of training data (\ieours weak supervision). This is an interesting direction for further research, and we leave it for future exploration.

\begin{table}[t]
\centering
\resizebox{1.0\textwidth}{!}{%
\begin{tabular}{cccccccccccccccc}
\Xhline{2\arrayrulewidth}
Settings & Methods & \textbf{mIoU(\%)} & \textit{ceil.} & \textit{floor} & \textit{wall} & \textit{beam} & \textit{col.} & \textit{win.} & \textit{door} & \textit{chair} & \textit{table} & \textit{book.} & \textit{sofa} & \textit{board} & \textit{clutter} \\ 
\hline
100\% & \nickname{} & 63.73 & 92.76 & 96.92 & \textbf{81.84} & 0.00 & 25.93 & 50.53 & 65.88 & 79.52 & 85.31 & 55.66 & \textbf{72.51} & 65.78 & 55.85 \\
10\% & \nickname{} & \textbf{64.67} & \textbf{93.04} & \textbf{97.45} & 81.55 & 0.00 & \textbf{28.01} & 55.77 & 68.68 & \textbf{80.11} & \textbf{87.67} & 55.25 & 72.31 & 63.91 & \textbf{57.02} \\
1\% & \nickname{} & 63.65 & 92.03 & 96.41 & 81.32 & 0.00 & 21.42 & 53.71 & \textbf{73.17} & 77.80 & 85.95 & 56.72 & 69.91 & \textbf{66.57} & 52.49 \\
0.1\% & \nickname{} & 61.41 & 91.72 & 95.63 & 78.71 & 0.00 & 24.23 & \textbf{55.89} & 63.14 & 70.50 & 83.13 & \textbf{60.67} & 67.82 & 56.14 & 50.63 \\
0.01\% & \nickname{} & 45.30 & 89.16 & 93.49 & 71.28 & 0.00 & 4.14 & 34.67 & 41.02 & 54.88 & 66.85 & 25.68 & 55.37 & 12.80 & 39.57 \\
\Xhline{2\arrayrulewidth}
\end{tabular}%
}
\caption{Quantitative results achieved by our \nickname{} on \textit{Area-5} of S3DIS under different amounts of labeled points.}
\label{tab:annotation}
\end{table}

\begin{table}[t]
\centering
\resizebox{0.75\textwidth}{!}{%
\begin{tabular}{r|ccc}
\Xhline{2.0\arrayrulewidth}
 & SPVCNN \cite{e3d} & MinkowskiUnet \cite{4dMinkpwski} & \textbf{SQN (Ours)} \\ 
 \Xhline{2.0\arrayrulewidth}
 Random & 49.61 & 46.15 & \textbf{60.19} \\
Softmax Confidence \cite{wang2014new} & 51.05 & 45.45 & \textbf{57.24} \\
Softmax Margin \cite{wang2014new} & 50.80 & 44.33 & \textbf{57.94} \\
Softmax Entropy \cite{wang2014new} & 50.35 & 49.99 & \textbf{57.98} \\
MC Dropout \cite{gal2016dropout} & 50.39 & 49.94 & \textbf{58.30} \\
ReDAL \cite{wu2021redal} & 50.89 & 47.88 & \textbf{54.24} \\
\Xhline{2.0\arrayrulewidth}
\end{tabular}%
}
\caption{Quantitative results achieved by different methods on the region-wise labeled S3DIS dataset.}
\label{tab:region_wise}
\end{table}

\smallskip\noindent\textbf{(5) Extension to Region-wise Annotated Data.} Beyond evaluating our method on randomly point-wise annotated datasets, we also extend our \nickname{} on the region-wise sparsely labeled S3DIS dataset. Following \cite{wu2021redal}, point clouds are firstly grouped into regions by unsupervised over-segmentation methods \cite{vcss}, and then a sparse number of regions are manually annotated through various active learning strategies \cite{wang2014new,wu2021redal,gal2016dropout}. As shown in Table \ref{tab:region_wise}, our SQN can consistently achieve better results than vanilla SPVCNN \cite{e3d} and MinkowskiNet \cite{4dMinkpwski} under the same supervision signal (10 iterations of active selection), regardless of the active learning strategy used. This is likely because the SparseConv based methods \cite{e3d,4dMinkpwski} usually have larger models and more trainable parameters compared with our point-based lightweight SQN, thus naturally exhibiting a stronger demand and dependence for more supervision signals. On the other hand, this result  further validates the effectiveness and superiority of our SQN under weak supervision.

\section{Conclusion}
In this paper, we propose \nickname{}, a conceptually simple and elegant framework to learn the semantics of large-scale point clouds, with as few as 0.1\% supplied labels for training. We first point out the redundancy of dense 3D annotations through extensive experiments, and then propose an effective semantic query framework based on the assumption of semantic similarity of neighboring points in 3D space. The proposed \nickname{} simply follows the concept of wider label propagation, but shows great potential for weakly-supervised semantic segmentation of large-scale point clouds. It would be interesting to extend this method for weakly-supervised instance segmentation, panoptic segmentation, and interactive annotation based on active learning.

\clearpage
%
%
\bibliographystyle{splncs04}
\bibliography{egbib.bib}

\clearpage
\section*{Appendix}

\section{Details of Sparse Annotation Tool}

\smallskip\noindent\textbf{(1) Annotation Pipeline.} As mentioned in Section \ref{sec:benchmark}, we develop a user-friendly annotation pipeline based on the off-the-shelf software. Note that, this tool is important to justify the feasibility/suitability of the low-cost random sparse annotation scheme, as most existing methods have directly overlooked this or taken it for granted that such tool is available. Here, we provide more detailed information on the pipeline. Specifically, given large-scale raw point clouds, the sparse annotation pipeline could be generally divided into the following steps:
\begin{enumerate}
\setlength{\itemsep}{0pt}
\setlength{\parsep}{0pt}
\setlength{\parskip}{0pt}
    \item Load the raw point clouds;
    \item Random downsample to a specified ratio (\textit{e.g.,} 0.1\%);
    \item Increase the point size of the downsampled points;
    \item Visualize the original point cloud and the down-sampled point cloud simultaneously;
    \item Annotate downsampled points in polygonal edition mode;
    \item Refine point labels.
\end{enumerate}

We also provide a video illustrating the annotation pipeline, which can be viewed at \url{https://youtu.be/N0UAeY31msY}.

\smallskip\noindent\textbf{(2) The Number of Annotated Points on 7 Datasets.} Considering that the existing large-scale point cloud datasets usually have millions of points, and typically have relatively high density, we therefore follow \cite{thomas2019kpconv,hu2019randla} to perform grid downsampling of raw point clouds at the beginning, and then execute the random based annotation steps in practice. Note that, all experiments of our SQN on the seven public datasets  follow this setting. As shown in Table \ref{tab:grid_sampling}, the grid downsampling at the beginning can significantly reduce the number of raw points. Taking the Semantic3D dataset which has high density as an example, the total number of points after grid downsampling becomes 1/50 of the original point clouds. Following the 0.1\% sparse annotation pipeline in our SQN, the total number of annotated points is only 78100, which is an approximately 0.002\% of the total raw points. To avoid confusion, we still report the 0.1\% labeling ratio in the main paper to keep consistency (\textit{i.e.,} the number of annotated points after grid downsampling / the total number of points after grid downsampling). Importantly, this is significantly different from 1T1C \cite{liu2021one} and cannot be directly compared, which calculates its labeling ratio by using the number of labeled instances divided by the total number of points, so as to achieve an over-exaggerated labeling ratio.

\begin{table}[t]
\centering
\resizebox{0.7\textwidth}{!}{%
\begin{tabular}{rcccc}
\Xhline{2\arrayrulewidth}
 & Grid size & Raw pts & Grid sampled pts & Anno. pts (0.1\%) \\ 
\Xhline{2\arrayrulewidth}
S3DIS \cite{2D-3D-S} & 0.04  & 273M  & 18.6M & 18,600 \\
Semantic3D \cite{Semantic3D} & 0.06  & 4000M  & 78.1M & 78,100  \\
ScanNet \cite{varney2020dales} & 0.04  & 242M  &  60.2M  & 60,200\\ 
SemanticKITTI \cite{behley2019semantickitti} & 0.06  & 5299M  & 3401M  & 3.4M \\
DALES \cite{varney2020dales} & 0.32  & 505M  & 211M  & 211,000\\
SensatUrban \cite{hu2020towards} & 0.2 & 2847M  & 221M  & 221,000\\
Toronto3D \cite{Toronto3D} & 0.04  & 78.3M  & 24.3M & 24,300 \\
\Xhline{2\arrayrulewidth}
\end{tabular}%
}
\caption{A comparison of the total number of points (M: Million) before and after grid sampling for seven public datasets. The grid size and the number of actual annotated points under our 0.1\% supervision setting are also reported.}
\label{tab:grid_sampling}
\end{table}

\smallskip\noindent\textbf{(3) Annotation Cost.} The sparse annotation scheme used in our \nickname{} can greatly reduce the annotation cost in practice, especially for extremely large-scale 3D point clouds with billions of points. Taking 0.1\% sparse point annotation as an example, with the developed CloudComapre-based labelling tool, a professional annotator can finish the annotation of the whole SensatUrban \cite{hu2020towards} dataset within \textbf{16} hours. By comparison, the original dense point-wise labeling costs \textbf{600} person-hours. Primarily, \textbf{this is because the random annotation based pipeline offers great error tolerance to avoid annotating boundary areas} (as only a very small number of points fall on the boundary), hence advanced functions such as polygonal edition can be freely and flexibly use, finally improve the productivity. In the traditional dense labeling pipeline, annotators are usually required to rotate and zoom back and forth to accurately separate the boundary areas, which consumes most of the labeling time. However, the random annotation based pipeline used in our developed tool can greatly reduce such time-consuming labelling of boundary areas, hence greatly reducing the overall annotation cost. Note that, the annotation cost (\textit{i.e.,} the total annotation time) could be further reduced if more advanced annotation software such as QTModeler\footnote{https://appliedimagery.com/} is used, where the user interface is more friendly.

\section{Implementation Tricks}
\smallskip\noindent\textbf{(1) Data augmentation.} We follow \cite{xu2020weakly} to apply different data augmentation techniques on the input point clouds during training, including random flipping, random rotation, and random noise.

\smallskip\noindent\textbf{(2) Re-training with generated pseudo labels.} We observe that different datasets (\egours S3DIS \cite{2D-3D-S} \textit{vs}. Semantic3D \cite{Semantic3D}) have significantly different number of total points (273 million \textit{vs}. 4000 million points). Therefore, the actual number of annotated points under our weak supervision setting (0.1\%) are also different (18600 \textit{vs}. 78100, as reported in Table \ref{tab:grid_sampling}). In the experiment, for the relatively small-scale S3DIS dataset which has extremely sparse supervision signals, we empirically find that retraining a new model with the generated pseudo labels can further increase the final segmentation performance. In particular, we firstly train our \nickname{} with the limited annotated 0.1\% points, and then infer the semantics of the entire training set. These estimated semantics are regarded as pseudo labels. After that, we retrain a new model of our \nickname{} from scratch with the generated pseudo labels. This retraining trick is able to fully utilize the extremely limited but valuable supervision signals. However, for large-scale datasets including Semantic3D \cite{Semantic3D}, SensatUrban \cite{hu2020towards}, SemanticKITTI \cite{behley2019semantickitti}, DALES \cite{varney2020dales} in Section \ref{subsec:Large-Scale_3D_Benchmarks}, our \nickname{} can achieve satisfactory results trained with 0.1\% annotated points, while the retraining trick does not noticeably improve the performance. Advanced techniques such as pseudo label refining \cite{zhang2021flexmatch} will be further explored in future work.

\section{Video Illustration}
We also provide a video illustrating the performance achieved by proposed \nickname{}, which can be viewed at \url{https://youtu.be/Q6wICSRRw3s}.

\section{Additional Ablation Results}

\smallskip\noindent\textbf{(1) Varying backbones of our SQN framework.} To further study the performance of our SQN framework with different backbones, we further implement our SQN based on the representative voxel-based baseline MinkowskiNet \cite{4dMinkpwski}. Specifically, we follow the implementation provided in \cite{e3d}, and the point local feature extractor, in this case, includes 4 encoding layers, each containing a 3D convolution block (kernel size and stride are set as 2) and 2 residual blocks (kernel size and stride are set as 3 and 1, respectively). Additionally, the feature query network gathers feature vectors from multi-level feature volumes through trilinear interpolation, and then simply concatenated and sent to MLPs (256-128-96) for semantic prediction.

\begin{table}[t]
\centering
\resizebox{\textwidth}{!}{%
\begin{tabular}{r|ccccccccccccccccccccc}
\Xhline{2.0\arrayrulewidth}
Methods & \rotatebox{90}{\textbf{mIoU(\%)}} & \rotatebox{90}{Params(M)} & \rotatebox{90}{\textit{road}} & \rotatebox{90}{\textit{sidewalk}} & \rotatebox{90}{\textit{parking}} & \rotatebox{90}{\textit{other-ground}} & \rotatebox{90}{\textit{building}} & \rotatebox{90}{\textit{car}} & \rotatebox{90}{\textit{truck}} & \rotatebox{90}{\textit{bicycle}} & \rotatebox{90}{\textit{motorcycle}} & \rotatebox{90}{\textit{other-vehicle}} & \rotatebox{90}{\textit{vegetation}} & \rotatebox{90}{\textit{trunk}} & \rotatebox{90}{\textit{terrain}} & \rotatebox{90}{\textit{person}} & \rotatebox{90}{\textit{bicyclist}} & \rotatebox{90}{\textit{motorcyclist}} & \rotatebox{90}{\textit{fence}} & \rotatebox{90}{\textit{pole}} & \rotatebox{90}{\textit{traffic-sign}} \\
\hline
    MinkUNet 0.1\% & 55.5  & 21.9  & 92.5  & 79.0  & 43.0  & 0.9   & 88.7  & 95.0  & 64.5  & 0.9   & 47.4  & 46.4  & 87.3  & 63.4  & 73.7  & 45.3  & 70.3  & 0.3   & 53.4  & 59.9  & 43.2  \\
    \textbf{SQN (MinkUNet)} 0.1\% & 55.8  & 8.8   & 91.5  & 78.0  & 41.1  & 0.9   & 88.5  & 94.9  & 66.8  & 5.7   & 43.2  & 43.6  & 88.2  & 64.0  & 75.5  & 49.5  & 66.3  & 0.0   & 55.9  & 61.1  & 45.2  \\
    MinkUNet 0.01\% & 43.2  & 21.9  & 89.3  & 74.8  & 32.1  & 0.0   & 87.6  & 92.4  & 25.8  & 0.0   & 24.8  & 20.1  & 87.1  & 56.4  & 73.2  & 9.6   & 15.9  & 0.0   & 55.1  & 48.2  & 28.3  \\
    \textbf{SQN (MinkUNet)} 0.01\% & 50.0  & 8.8   & 89.7  & 75.6  & 31.9  & 0.2   & 87.6  & 93.5  & 47.2  & 0.2   & 35.6  & 31.6  & 88.2  & 58.0  & 76.0  & 33.8  & 59.1  & 0.0   & 52.9  & 52.1  & 36.4  \\
\Xhline{2.0\arrayrulewidth}
    \end{tabular}
    }
\caption{Quantitative results achieved by our SQN (SPVNAS) on the validation set of the SemanticKITTI dataset under different weak supervision settings.}
  \label{tab:SPVNAS}%
\end{table}%

The experimental results achieved by our SQN and baseline networks on the SemanticKITTI dataset under different weak supervision settings are shown in Table \ref{tab:SPVNAS}. We can see that our SQN achieves comparable performance with the baseline under 0.1\% settings, primarily because the supervision signal is still sufficient at this time, considering the large scale of the dataset. However, we can clearly observe that our SQN outperforms the baseline by a large margin (6.8\% improvement in mIoU scores) when there are only 0.01\% points are annotated. This further demonstrates the effectiveness of our semantic query framework.

\smallskip\noindent\textbf{(2) Comparison of our SQN with other feature propagation layers.} Here, we explicitly discuss the key differences of our SQN and other general feature propagation layers used in \cite{qi2017pointnet++,hu2019randla}. To clarify, general feature propagation layers are primarily used to recover the full spatial resolution for dense segmentation given \textbf{full supervision}, while SQN queries parallelly and hierarchically at multiple spatial resolutions, aiming to effectively propagate the limited signals to a much wider context given \textbf{weak supervision}. Further, we compare our SQN with two general feature propagation layers by conducting ablative experiments on three datasets. Specifically, we keep the feature encoder, labeling ratio, and experimental settings unchanged, and only replace the semantic query decoder as the vanilla feature propagation layers used in PointNet++ and RandLA-Net. The experimental results are shown in Table \ref{tab:propogation}. It can be seen that our SQN achieves consistently better results compared with two general feature propagation layers, primarily due to the usage of parallelly and hierarchically semantic queries in different encoding layers, enabling the limited and sparse supervision signal to be back-propagated to a much wider context. By contrast, both the feature propagation layers used in RandLA-Net and PointNet++ can only propagate the sparse label layer by layer, with relatively limited receptive fields, hence achieving inferior performance in the weak supervision settings.

\begin{table}[ht]
\centering
    \resizebox{0.75\textwidth}{!}{%
\begin{tabular}{rccc}
\Xhline{2.0\arrayrulewidth}
 & S3DIS & ScanNet (val) & Toronto3D \\ 
\Xhline{2.0\arrayrulewidth}
FP layers in RandLA-Net \cite{hu2019randla} & 52.9 & 51.2  & 71.7  \\
FP layers in PointNet++ \cite{qi2017pointnet++} & 53.4  & 52.7  & 72.4 \\
\textbf{SQN (Ours)} & 61.4 & 58.4 & 77.7 \\ 
\Xhline{2.0\arrayrulewidth}
\end{tabular}%
}
\caption{Ablative experiments on different feature propagation (FP) layers.}
\label{tab:propogation}
\end{table}

\smallskip\noindent\textbf{(2) Detailed Results on Varying Annotated Points.} In Section \ref{Sec:ablation}, we evaluate the sensitivity of the proposed \nickname{} to different randomly annotated points. Here, we provide detailed experimental results on Table \ref{tab:Sensitivity}. It can be seen that the major performance variations are in minor categories such as \textit{door}, \textit{sofa}, and \textit{board}, indicating that the underrepresented categories are more sensitive to our weakly-supervised settings, \textit{i.e.,} 0.1\% random annotated point labels. This is not surprising because such imbalanced distribution issue also widely exists in fully-supervised methods.

\begin{table*}[t]
\centering
\resizebox{\textwidth}{!}{%
\begin{tabular}{rccccccccccccccc}
\Xhline{2\arrayrulewidth}
& OA(\%) & \textbf{mIoU(\%)} & \textit{ceil.} & \textit{floor} & \textit{wall} & \textit{beam} & \textit{col.} & \textit{wind.} & \textit{door} & \textit{table} & \textit{chair} & \textit{sofa} & \textit{book.} & \textit{board} & \textit{clut.} \\
\hline
Iter1 & \textbf{86.53} & 60.97 & \textbf{92.33} & 96.70 & \textbf{78.99} & 0.00 & 25.01 & \textbf{56.76} & 58.99 & \textbf{74.22} & 79.06 & 58.41 & 67.73 & 53.29 & 51.08 \\
Iter2 & 85.63 & 59.24 & 91.72 & \textbf{97.01} & 77.35 & 0.00 & 20.10 & 53.55 & \textbf{65.28} & 71.63 & \textbf{83.61} & 51.44 & 65.57 & 43.37 & 49.49 \\
Iter3 & 86.39 & 60.93 & 91.96 & 96.02 & 78.88 & 0.00 & \textbf{25.31} & 55.80 & 63.43 & 70.71 & 82.80 & 51.18 & \textbf{68.39} & 56.53 & 51.05 \\
Iter4 & 86.32 & 59.40 & 92.22 & 96.07 & 78.85 & 0.00 & 19.00 & 50.10 & 65.19 & 68.37 & 83.27 & 49.79 & 67.09 & 51.33 & 50.89 \\
Iter5 & 86.40 & \textbf{61.56} & 91.88 & 95.97 & 78.89 & 0.00 & 24.95 & 55.88 & 63.73 & 70.75 & 83.20 & \textbf{59.29} & 68.25 & \textbf{56.37} & \textbf{51.13} \\
\textbf{Average} & 86.25 & 60.42 & 92.02 & 96.35 & 78.59 & 0.00 & 22.87 & 54.42 & 63.32 & 71.14 & 82.39 & 54.02 & 67.41 & 52.18 & 50.73 \\
\textbf{STD} & 0.32 & 0.93 & 0.22 & 0.42 & 0.62 & 0.00 & 2.74 & 2.41 & 2.29 & 1.88 & 1.68 & 3.99 & 1.03 & 4.82 & 0.62 \\
\Xhline{2\arrayrulewidth}
\end{tabular}%
}
\caption{Sensitivity analysis of the proposed \nickname{} on the S3DIS dataset (\textit{Area 5}) by running 5 times. Overall Accuracy (OA, \%), mean IoU (mIoU, \%), and per-class IoU (\%) are reported. Bold represents the best result.}
\label{tab:Sensitivity}
\end{table*}

\section{Additional Discussion}

\begin{figure}[t]
	\begin{center}
		\includegraphics[width=0.75\textwidth]{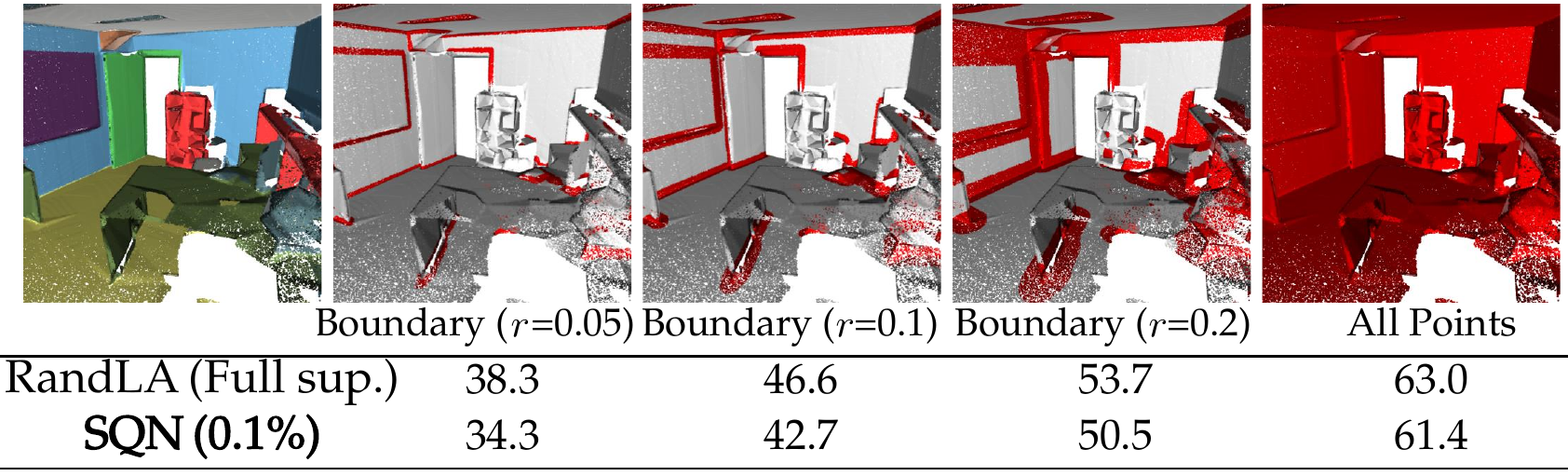}
	\end{center}
    \caption{Quantitative comparison of RandLA-Net and the proposed \nickname{} on the boundary areas.}
	\label{fig: boundary}
\end{figure}

\begin{figure}[t]
	\begin{center}
		\includegraphics[width=0.9\textwidth]{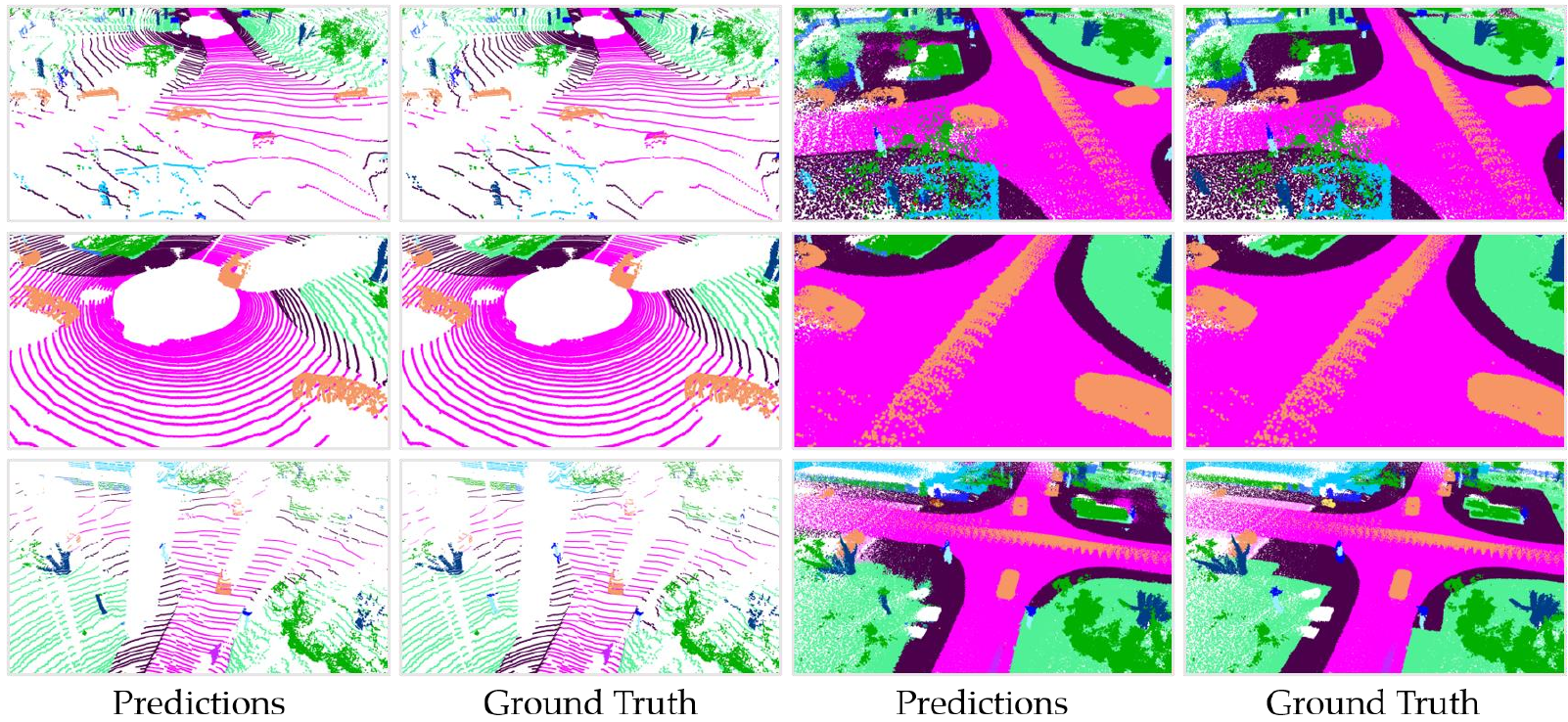}
	\end{center}
    \caption{Qualitative performance achieved by \nickname{} on the SemanticKITTI dataset when trained on partial point clouds.}
	\label{fig: inference}
\end{figure}

\smallskip\noindent\textbf{(1) Performance on Boundary Areas.} We further evaluate the segmentation performance of our \nickname{} on the boundary points, since the assumption about the consistency of neighborhood semantics may not hold at the boundary areas with different semantics. Specifically, we first define the boundary points as follows: if the queried spherical neighboring points within a radius $r$ have different semantics, the query point is regarded as on the boundary (red points in the Figure \ref{fig: boundary}). Not surprisingly, given a smaller $r$, the performance drops significantly for both RandLA-Net (full supervision) and SQN (0.1\%), showing that it is still a common issue for existing methods. We will leave this issue for future exploration.

\smallskip\noindent\textbf{(2) Flexibility of \nickname{}.} Thanks to the flexibility of the SQN framework, the proposed method should be able to take any point in the space as input and infer its semantic label through query and interpolation (even if that point itself does not exist in the point clouds). To further validate this, we tried to train our SQN on the partial point clouds (\ieours raw point clouds), but test on the aggregated point clouds based on the SemanticKITTI dataset. The qualitative results are shown in Figure \ref{fig: inference}. It can be seen that the proposed SQN can still achieve satisfactory performance on the complete point clouds, even though our model is only trained with partial and incomplete point clouds. Considering that point clouds are irregularly sampled points from the continuous surface, it would be interesting to further explore the continuous semantic surface learning based on our framework.

\smallskip\noindent\textbf{(3) Potential Negative Societal Impact.} Our work aims to achieve label-efficient learning of large-scale 3D point clouds, which could potentially be used in autonomous driving or robotics systems. It is targeted for semantic segmentation of 3D point clouds with weak supervision, hence there is no known society negative impact. However, the robustness, security, safety issues should be further checked before being applied in real-world data.

\smallskip\noindent\textbf{(4) Limitation and Future Work.} Our \nickname{} is intuitive simple yet effective. Extensive experiments have demonstrated the superiority on large-scale datasets. However, it still relies on human annotations, albeit extremely sparse. Ideally, the 3D semantics can be automatically discovered from raw point clouds. We leave this unsupervised learning of 3D semantic segmentation for our future exploration.

\section{Additional Experimental Results}

\smallskip\noindent\textbf{(1) Detailed Results of Fully Supervised Baselines under Sparse Annotations.} As mentioned in Section \ref{sec:benchmark}, we evaluate several baseline methods under different forms of weak supervision. Here, we further provide the detailed benchmarking results on Table \ref{tab:S3DIS_baselines}, with per-class IoU scores reported. Note that, this table is corresponding to Figure \ref{fig: critical_point} in the main paper.

\begin{table*}[thb]
\centering
\resizebox{0.95\textwidth}{!}{%
\begin{tabular}{c|rcccccccccccccc}
\Xhline{2\arrayrulewidth}
 Settings & Methods & \textbf{mIoU(\%)} & \textit{ceil.} & \textit{floor} & \textit{wall} & \textit{beam} & \textit{col.} & \textit{win.} & \textit{door} & \textit{chair} & \textit{table} & \textit{book.} & \textit{sofa} & \textit{board} & \textit{clutter} \\ 
\hline
\multirow{3}{*}{\begin{tabular}[c]{@{}c@{}}100\%\\ (Full \\ supervision)\end{tabular}} 
 & PointNet \cite{qi2017pointnet} & 39.15 & 89.65 & 93.37 & 70.32 & 0.00 & 0.85 & 36.22 & 3.03 & 57.29 & 44.40 & 0.02 & 56.21 & 19.65 & 37.95 \\
 & PointNet++ \cite{qi2017pointnet++} &52.36 & 88.84 & 90.88 & 75.83 & 0.18 & 10.47 & 43.57 & 13.86 & 71.90 & 82.81 & 35.71 & 67.28 & 51.60 & 47.80 \\
 & RandLA-Net \cite{hu2019randla} & 63.75 & 92.19 & 97.67 & 81.12 & 0.00 & 20.22 & 61.02 & 41.49 & 78.53 & 88.04 & 70.65 & 74.21 & 70.65 & 53.01 \\
 \hline
\multirow{3}{*}{\begin{tabular}[c]{@{}c@{}}10\%\\ (Random)\end{tabular}} 
 & PointNet \cite{qi2017pointnet} & 38.41 & 88.65 & 94.20 & 71.11 & 0.00 & 0.15 & 27.16 & 4.28 & 58.34 & 45.28 & 0.05 & 54.58 & 18.89 & 36.63 \\
 & PointNet++ \cite{qi2017pointnet++} &  52.34 & 86.67 & 90.68 & 76.37 & 0.00 & 10.63 & 43.76 & 20.14 & 70.37 & 83.34 & 40.97 & 68.00 & 41.88 & 47.64  \\
 & RandLA-Net \cite{hu2019randla} & 61.87 & 91.87 & 97.58 & 79.71 & 0.00 & 19.24 & 60.76 & 39.36 & 77.06 & 86.44 & 61.77 & 70.63 & 67.50 & 52.34 \\
\hline
\multirow{3}{*}{\begin{tabular}[c]{@{}c@{}}1\%\\ (Random)\end{tabular}} 
 & PointNet \cite{qi2017pointnet} & 37.23 & 88.93 & 94.90 & 68.94 & 0.00 & 0.18 & 21.76 & 3.22 & 56.44 & 44.29 & 0.06 & 52.08 & 17.78 & 35.47 \\
 & PointNet++ \cite{qi2017pointnet++} & 48.61 & 87.65 & 89.39 & 73.98 & 0.01 & 7.05 & 39.15 & 12.98 & 66.28 & 73.94 & 28.97 & 66.87 & 40.13 & 45.56 \\
 & RandLA-Net \cite{hu2019randla} & 59.13 & 90.86 & 96.96 & 78.34 & 0.00 & 16.40 & 60.33 & 25.73 & 75.30 & 83.05 & 59.10 & 69.00 & 64.84 & 48.73 \\
\hline
\multirow{3}{*}{\begin{tabular}[c]{@{}c@{}}0.1\%\\ (Random)\end{tabular}} 
& PointNet \cite{qi2017pointnet} & 33.26 & 83.48 & 89.40 & 61.66 & 0.00 & 0.01 & 20.85 & 3.82 & 48.57 & 31.80 & 3.77 & 41.08 & 21.99 & 25.96 \\
& PointNet++ \cite{qi2017pointnet++} & 42.57 & 85.43 & 88.76 & 69.87 & 0.00 & 1.00 & 24.61 & 7.30 & 57.72 & 66.28 & 24.90 & 58.80 & 30.89 & 37.82 \\
& RandLA-Net \cite{hu2019randla} & 52.90 & 89.90 & 95.90 & 75.28 & 0.00 & 7.46 & 52.38 & 26.48 & 62.19 & 74.48 & 49.10 & 60.15 & 49.26 & 45.08 \\
\hline
\multirow{3}{*}{\begin{tabular}[c]{@{}c@{}}0.01\%\\ (Random)\end{tabular}} 
& PointNet \cite{qi2017pointnet} & 21.28 & 72.13 & 81.79 & 53.48 & 0.00 & 0.00 & 7.03 & 4.66 & 24.40 & 8.39 & 0.00 & 8.51 & 0.00 & 16.30 \\
& PointNet++ \cite{qi2017pointnet++} & 33.53 & 77.84 & 83.87 & 67.09 & 0.23 & 3.89 & 34.83 & 16.60 & 41.49 & 30.65 & 0.79 & 39.23 & 13.81 & 25.50 \\
& RandLA-Net \cite{hu2019randla} & 33.16 & 85.15 & 89.20 & 61.54 & 0.00 & 3.66 & 13.17 & 9.11 & 29.15 & 42.29 & 6.52 & 46.78 & 16.86 & 27.72 \\
\Xhline{2\arrayrulewidth}
\end{tabular}%
}
\caption{Detailed benchmark results of three baselines in the \textit{Area-5} of the S3DIS \cite{2D-3D-S} dataset. Different amount of points are randomly annotated for weak supervision.}
\label{tab:S3DIS_baselines}
\end{table*}

\smallskip\noindent\textbf{(2) Additional Results on S3DIS.} In Section \ref{subsec:Comparison_with_SOTA}, we provide the quantitative results achieved on the \textit{Area-5} subset of the S3DIS dataset. Here, we further report the detailed 6-fold cross-validation results achieved by our \nickname{} and other baselines on this dataset in Table \ref{tab:6-fold-cv}.


\begin{table*}[thb]
\centering
\resizebox{\textwidth}{!}{%
\begin{tabular}{c|rcccccccccccccccc}
\Xhline{2\arrayrulewidth}
& Methods & OA(\%) & mAcc(\%)& \textbf{mIoU(\%)} & \textit{ceil.} & \textit{floor} & \textit{wall} & \textit{beam} & \textit{col.} & \textit{wind.} & \textit{door} & \textit{table} & \textit{chair} & \textit{sofa} & \textit{book.} & \textit{board} & \textit{clut.} \\ 
\hline
\multirow{11}{*}{\begin{tabular}[c]{@{}c@{}}Full\\ supervision\end{tabular}} 
& PointNet \cite{qi2017pointnet} & 78.6 & 66.2 & 47.6 & 88.0 & 88.7 & 69.3 & 42.4 & 23.1 & 47.5 & 51.6 & 54.1 & 42.0 & 9.6 & 38.2 & 29.4 & 35.2 \\
& RSNet \cite{RSNet} & - & 66.5 & 56.5 & 92.5 & 92.8 & 78.6 & 32.8 & 34.4 & 51.6 & 68.1 & 59.7 & 60.1 & 16.4 & 50.2 & 44.9 & 52.0 \\
& 3P-RNN \cite{3PRNN} & 86.9 & - & 56.3 & 92.9 & 93.8 & 73.1 & 42.5 & 25.9 & 47.6 & 59.2 & 60.4 & 66.7 & 24.8 & 57.0 & 36.7 & 51.6 \\
& SPG \cite{landrieu2018large}& 86.4 & 73.0 & 62.1 & 89.9 & 95.1 & 76.4 & 62.8 & 47.1 & 55.3 & 68.4 & \uline{73.5} & 69.2 & 63.2 & 45.9 & 8.7 & 52.9 \\
& PointCNN \cite{li2018pointcnn} & 88.1 & 75.6 & 65.4 & 94.8 & 97.3 & 75.8 & 63.3 & 51.7 & 58.4 & 57.2 & 71.6 & 69.1 & 39.1 & 61.2 & 52.2 & 58.6 \\ 
& PointWeb \cite{pointweb} & 87.3 & 76.2 & 66.7 & 93.5 & 94.2 & 80.8 & 52.4 & 41.3 & 64.9 & 68.1 & 71.4 & 67.1 & 50.3 & 62.7 & 62.2 & 58.5  \\
& ShellNet \cite{zhang2019shellnet} & 87.1  & - & 66.8 & 90.2 & 93.6 & 79.9 & 60.4 & 44.1 & 64.9 & 52.9 & 71.6 & \uline{84.7} & 53.8 & 64.6 & 48.6 & 59.4 \\
& PointASNL \cite{yan2020pointasnl} & \uline{88.8}  & 79.0 & 68.7 & \uline{95.3} & \uline{97.9} & 81.9 & 47.0 & 48.0 & \uline{67.3} & 70.5 & 71.3 & 77.8 & 50.7 & 60.4 & 63.0 & \uline{62.8} \\
& KPConv (\textit{rigid}) \cite{thomas2019kpconv} & - & 78.1 & 69.6 & 93.7 & 92.0 & 82.5 & 62.5 & 49.5  & 65.7 & \uline{77.3} & 57.8 & 64.0 & 68.8 & 71.7 & 60.1 & 59.6 \\
& KPConv (\textit{deform}) \cite{thomas2019kpconv} &- & 79.1 &\uline{70.6} &93.6 &92.4 & \uline{83.1} & \uline{63.9} & \uline{54.3} & 66.1 & 76.6 &57.8 &64.0 & \uline{69.3} & \uline{74.9} &61.3 & 60.3 \\
& RandLA-Net \cite{hu2019randla} & 88.0 & \uline{82.0} & 70.0 & 93.1 & 96.1 & 80.6 & 62.4 & 48.0 & 64.4 & 69.4 & 69.4 & 76.4 & 60.0 & 64.2 & \uline{65.9} & 60.1 \\
\hline
\multirow{1}{*}{\begin{tabular}[c]{@{}c@{}}Weak sup.\end{tabular}} 
& \textbf{\nicknamenew{} (0.1\%)}  & \textbf{85.3} & \textbf{76.3} & \textbf{63.7} & \textbf{92.5} & \textbf{95.4} & \textbf{77.1} & \textbf{50.8} & \textbf{43.6} & \textbf{58.5} & \textbf{67.0} & \textbf{67.7} & \textbf{54.1} & \textbf{54.9} & \textbf{61.0} & \textbf{53.0} & \textbf{52.7} \\
 \Xhline{2\arrayrulewidth}
\end{tabular}%
}
\caption{Quantitative results of different approaches on S3DIS \cite{2D-3D-S} (\textit{6-fold cross-validation}). Overall Accuracy (OA, \%), mean class Accuracy (mAcc, \%), mean IoU (mIoU, \%), and per-class IoU (\%) are reported.}
\label{tab:6-fold-cv}
\end{table*}

\smallskip\noindent\textbf{(3) Additional Results on ScanNet.} The ScanNet \cite{scannet} dataset consists of 1613 indoor scans (1201 for training, 312 for validation, and 100 for online testing). It has nearly 242 million points sampled from the densely reconstructed 3D meshes. We provided the detailed per class IoU results on Table \ref{tab:scannet_full}.

\begin{table*}[thb]
\centering
\resizebox{\textwidth}{!}{%
\begin{tabular}{c|rccccccccccccccccccccc}
\Xhline{2.0\arrayrulewidth}
Settings & Method & \textbf{mIoU(\%)} & \textit{bath} & \textit{bed} & \textit{bkshf} & \textit{cab} & \textit{chair} & \textit{cntr} & \textit{curt} & \textit{desk} & \textit{door} & \textit{floor} & \textit{other} & \textit{pic} & \textit{fridg} & \textit{show} & \textit{sink} & \textit{sofa} & \textit{table} & \textit{toil} & \textit{wall} & \textit{wind} \\ \hline
\multirow{12}{*}{\begin{tabular}[c]{@{}c@{}}Full \\ supervision\end{tabular}}
 & ScanNet \cite{scannet} & 30.6 & 20.3 & 36.6 & 50.1 & 31.1 & 52.4 & 21.1 & 0.2 & 34.2 & 18.9 & 78.6 & 14.5 & 10.2 & 24.5 & 15.2 & 31.8 & 34.8 & 30.0 & 46.0 & 43.7 & 18.2 \\
 & PointNet++ \cite{qi2017pointnet++} & 33.9 & 58.4 & 47.8 & 45.8 & 25.6 & 36.0 & 25.0 & 24.7 & 27.8 & 26.1 & 67.7 & 18.3 & 11.7 & 21.2 & 14.5 & 36.4 & 34.6 & 23.2 & 54.8 & 52.3 & 25.2 \\
 & SPLATNET3D \cite{su2018splatnet} & 39.3 & 47.2 & 51.1 & 60.6 & 31.1 & 65.6 & 24.5 & 40.5 & 32.8 & 19.7 & 92.7 & 22.7 & 0.0 & 0.1 & 24.9 & 27.1 & 51.0 & 38.3 & 59.3 & 69.9 & 26.7 \\
 & Tangent-Conv \cite{tangentconv} & 43.8 & 43.7 & 64.6 & 47.4 & 36.9 & 64.5 & 35.3 & 25.8 & 28.2 & 27.9 & 91.8 & 29.8 & 14.7 & 28.3 & 29.4 & 48.7 & 56.2 & 42.7 & 61.9 & 63.3 & 35.2 \\
 & PointCNN \cite{li2018pointcnn} & 45.8 & 57.7 & 61.1 & 35.6 & 32.1 & 71.5 & 29.9 & 37.6 & 32.8 & 31.9 & 94.4 & 28.5 & 16.4 & 21.6 & 22.9 & 48.4 & 54.5 & 45.6 & 75.5 & 70.9 & 47.5 \\
 & PointConv \cite{wu2018pointconv} & 55.6 & 63.6 & 64.0 & 57.4 & 47.2 & 73.9 & 43.0 & 43.3 & 41.8 & 44.5 & 94.4 & 37.2 & 18.5 & 46.4 & 57.5 & 54.0 & 63.9 & 50.5 & 82.7 & 76.2 & 51.5 \\
 & SPH3D-GCN \cite{SPH3D} & 61.0 & \uline{85.8} & 77.2 & 48.9 & 53.2 & 79.2 & 40.4 & 64.3 & 57.0 & 50.7 & 93.5 & 41.4 & 4.6 & 51.0 & 70.2 & 60.2 & 70.5 & 54.9 & 85.9 & 77.3 & 53.4 \\
 & KPConv \cite{thomas2019kpconv} & 68.4 & 84.7 & 75.8 & 78.4 & 64.7 & 81.4 & 47.3 & \uline{77.2} & \uline{60.5} & 59.4 & 93.5 & 45.0 & 18.1 & 58.7 & 80.5 & 69.0 & 78.5 & 61.4 & 88.2 & 81.9 & 63.2 \\
 & SparseConvNet \cite{sparse} & \uline{72.5} & 64.7 & \uline{82.1} & \uline{84.6} & \uline{72.1} & \uline{86.9} & \uline{53.3} & 75.4 & 60.3 & \uline{61.4} & \uline{95.5} & \uline{57.2} & \uline{32.5} & \uline{71.0} & \uline{87.0} & \uline{72.4} & \uline{82.3} & \uline{62.8} & \uline{93.4} & \uline{86.5} & \uline{68.3} \\
 & SegGCN \cite{lei2020seggcn} & 58.9 & 83.3 & 73.1 & 53.9 & 51.4 & 78.9 & 44.8 & 46.7 & 57.3 & 48.4 & 93.6 & 39.6 & 6.1 & 50.1 & 50.7 & 59.4 & 70.0 & 56.3 & 87.4 & 77.1 & 49.3 \\
 & RandLA-Net \cite{hu2019randla} & 64.5 & 77.8 & 73.1 & 69.9 & 57.7 & 82.9 & 44.6 & 73.6 & 47.7 & 52.3 & 94.5 & 45.4 & 26.9 & 48.4 & 74.9 & 61.8 & 73.8 & 59.9 & 82.7 & 79.2 & 62.1 \\ \hline
\multirow{1}{*}{\begin{tabular}[c]{@{}c@{}}Weak sup.\end{tabular}} 
 & \textbf{Ours (0.1\%)} & \textbf{56.9} & \textbf{67.6} & \textbf{69.6} & \textbf{65.7} & \textbf{49.7} & \textbf{77.9} & \textbf{42.4} & \textbf{54.8} & \textbf{51.5} & \textbf{37.6} & \textbf{90.2} & \textbf{42.2} & \textbf{35.7} & \textbf{37.9} & \textbf{45.6} & \textbf{59.6} & \textbf{65.9} & \textbf{54.4} & \textbf{68.5} & \textbf{66.5} & \textbf{55.6} \\
\Xhline{2.0\arrayrulewidth}
\end{tabular}%
}
\caption{Quantitative results of different approaches on ScanNet (\textit{online test set}). Mean IoU (mIoU, \%), and per-class IoU (\%) scores are reported.  }
\label{tab:scannet_full}
\end{table*}

\begin{figure*}[h]
	\begin{center}
		\includegraphics[width=0.9\textwidth]{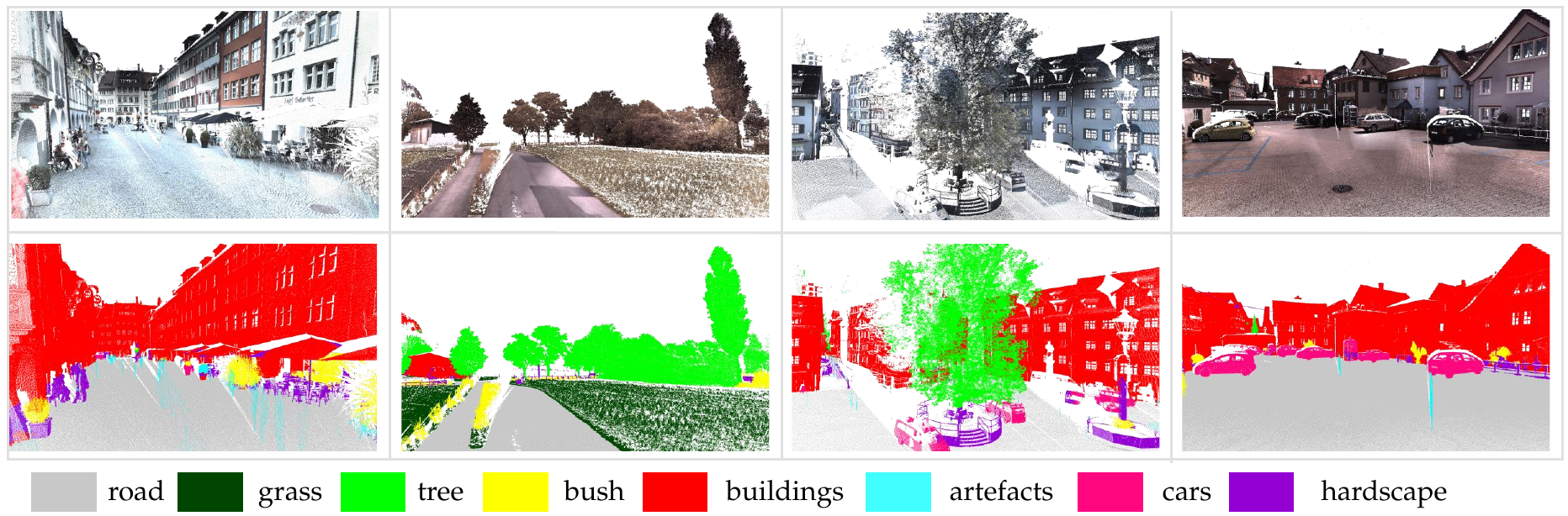}
	\end{center}
    \caption{Qualitative results achieved by our \nickname{} on the reduced-8 split of Semantic3D. Note that, the ground truth of the test set is not publicly available.}
	\label{fig: qualitative_semantic3d}
\end{figure*}

\smallskip\noindent\textbf{(4) Additional results on Semantic3D.} This dataset consists of 30 urban and rural street-scenarios (15 for training and 15 for online testing). There are 4 billion points in total acquired by the terrestrial laser. In particular, we also train our \nickname{} with only 0.01\% randomly annotated points, considering the extremely large amount of 3D points scanned. The detailed experimental results achieved on the \textit{Semantic8} and \textit{Reduced8} subset of the Semantic3D dataset are reported in Table \ref{tab:semantic-8} and Table \ref{tab:reduced-8}. In addition, we also show the qualitative results achieved by our \nickname{} on the \textit{Reduced-8} subset with 0.1\% labels in Fig \ref{fig: qualitative_semantic3d}.

\begin{table*}[ht]
\centering
\resizebox{0.9\textwidth}{!}{%
\begin{tabular}{c|rcccccccccc}
\Xhline{2.0\arrayrulewidth}
Settings & Methods & \textbf{mIoU(\%)} & OA(\%) & \textit{man-made.} & \textit{natural.} & \textit{high veg.} & \textit{low veg.} & \textit{buildings} & \textit{hard scape} & \textit{scanning art.} & \textit{cars} \\
\hline
\multirow{11}{*}{\begin{tabular}[c]{@{}c@{}}Full\\ supervision\end{tabular}} 
& TML-PC~\cite{montoya2014mind} & 39.1	& 74.5	& 80.4	& 66.1	& 42.3	& 41.2	& 64.7	& 12.4	& 0.0	& 5.8 \\
& TMLC-MS~\cite{hackel2016fast} & 49.4	& 85.0	& 91.1	& 69.5	& 32.8	& 21.6	& 87.6	& 25.9	& 11.3	& 55.3 \\
& PointNet++~\cite{qi2017pointnet++} & 63.1	& 85.7	& 81.9	& 78.1	& 64.3	& 51.7	& 75.9	& 36.4	& 43.7	& 72.6 \\
& EdgeConv~\cite{contreras2019edge} & 64.4	& 89.6	& 91.1	& 69.5	& 65.0	& 56.0	& 89.7	& 30.0	& 43.8	& 69.7 \\
& SnapNet~\cite{snapnet} & 67.4	& 91.0	& 89.6	& 79.5	& 74.8	& 56.1	& 90.9	& 36.5	& 34.3	& 77.2 \\
& PointGCR~\cite{ma2020global} & 69.5	& 92.1	& 93.8	& 80.0	& 64.4	& 66.4	& 93.2	& 39.2	& 34.3	& 85.3 \\
& RGNet~\cite{RGNet}	& 72.0	& 90.6  & 86.4	& 70.3	& 69.5  & 68.0	& 96.9	& 43.4	& 52.3	& 89.5 \\
& LCP ~\cite{boulch2020lightconvpoint} & 74.6	& 94.1	& 94.7	& 85.2	& 77.4	& 70.4	& 94.0	& 52.9	& 29.4	& 92.6 \\
& SPGraph~\cite{landrieu2018large} & 76.2	& 92.9	& 91.5	& 75.6	& \uline{78.3}	& 71.7	& 94.4	& \uline{56.8}	& 52.9	& 88.4 \\
& ConvPoint~\cite{conv_pts} & 76.5	& 93.4	& 92.1	& 80.6	& 76.0	& \uline{71.9}	& 95.6 & 47.3	& \uline{61.1}	& 87.7 \\
& RandLA-Net \cite{hu2019randla} & 75.8 & \uline{95.0} & \uline{97.4} & \uline{93.0} & 70.2 & 65.2 & 94.4 & 49.0 & 44.7 & 92.7 \\
& WreathProdNet \cite{wang2020equivariant} & \uline{77.1} & 94.6 & 95.2 & 87.1 & 75.3 & 67.1 & \uline{96.1} & 51.3 & 51.0 & \uline{93.4} \\
\hline
\multirow{2}{*}{\begin{tabular}[c]{@{}c@{}}Weak\\ supervision\end{tabular}} 
& \textbf{\nicknamenew{} (0.1\%)} & \textbf{72.3} & \textbf{94.8} & \textbf{97.9} & \textbf{93.2} & \textbf{65.5} & \textbf{63.4} & \textbf{94.9} & \textbf{44.9} & \textbf{47.4} & \textbf{70.9} \\
& \textbf{\nicknamenew{} (0.01\%)}  & 58.8 & 91.9 & 96.7 & 90.3 & 56.6 & 53.3 & 90.7 & 13.6 & 24.0 & 44.9 \\
\Xhline{2.0\arrayrulewidth}
\end{tabular}
}
\caption{Quantitative results of different approaches on Semantic3D (\textit{semantic-8}) \cite{Semantic3D}. This test consists of 2,091,952,018 points. The scores are obtained from the recent publications. Bold represents the best result in weakly-supervised methods, and underlined represents the best results in fully-supervised methods.}
\label{tab:semantic-8}
\end{table*}


\begin{table*}[th]
\centering
\resizebox{0.9\textwidth}{!}{%
\begin{tabular}{c|rcccccccccc}
\Xhline{2.0\arrayrulewidth}
Settings & Methods & \textbf{mIoU(\%)} & OA(\%) & \textit{man-made.} & \textit{natural.} & \textit{high veg.} & \textit{low veg.} & \textit{buildings} & \textit{hard scape} & \textit{scanning art.} & \textit{cars} \\
\hline
\multirow{10}{*}{\begin{tabular}[c]{@{}c@{}}Full\\ supervision\end{tabular}} 
& SnapNet\_ \cite{snapnet} & 59.1 & 88.6 & 82.0 & 77.3 & 79.7 & 22.9 & 91.1 & 18.4 & 37.3 & 64.4 \\
& SEGCloud \cite{tchapmi2017segcloud} & 61.3 & 88.1 & 83.9 & 66.0 & 86.0 & 40.5 & 91.1 & 30.9 & 27.5 & 64.3 \\
&RF\_MSSF \cite{RF_MSSF} & 62.7 & 90.3 & 87.6 & 80.3 & 81.8 & 36.4 & 92.2 & 24.1 & 42.6 & 56.6 \\
&MSDeepVoxNet \cite{msdeepvoxnet} & 65.3 & 88.4 & 83.0 & 67.2 & 83.8 & 36.7 & 92.4 & 31.3 & 50.0 & 78.2 \\
&ShellNet \cite{zhang2019shellnet} & 69.3 & 93.2 & 96.3 & 90.4 & 83.9 & 41.0 & 94.2 & 34.7 & 43.9 & 70.2 \\
 & GACNet \cite{GACNet} & 70.8 & 91.9 & 86.4 & 77.7 & \uline{88.5} & \uline{60.6} & 94.2 & 37.3 & 43.5 & 77.8 \\
 & SPG \cite{landrieu2018large} & 73.2 & 94.0 & 97.4 & 92.6 & 87.9 & 44.0 & 83.2 & 31.0 & 63.5 & 76.2 \\
 & KPConv \cite{thomas2019kpconv} & 74.6 & 92.9 & 90.9 & 82.2 & 84.2 & 47.9 & 94.9 & 40.0 & \uline{77.3} & \uline{79.9} \\
 & RGNet \cite{RGNet} & 74.7 & 94.5 & \uline{97.5} & \uline{93.0} & 88.1 & 48.1 & 94.6 & 36.2 & 72.0 & 68.0 \\
 & RandLA-Net \cite{hu2019randla} & \uline{77.4} & \uline{94.8} & 95.6 & 91.4 & 86.6 & 51.5 & \uline{95.7} & \uline{51.5} & 69.8 & 76.8 \\
\hline
\multirow{2}{*}{\begin{tabular}[c]{@{}c@{}}Weak\\ supervision\end{tabular}} 
& \textbf{\nicknamenew{} (0.1\%)} & \textbf{74.7} & \textbf{93.7} & \textbf{97.1} & \textbf{90.8} & \textbf{84.7} & \textbf{48.5} & \textbf{93.9} & \textbf{37.4} & \textbf{71.0} & \textbf{74.5} \\
& \textbf{\nicknamenew{} (0.01\%)}  &65.6 & 90.3 & 96.6 & 87.5 & 80.6 & 37.1 & 88.5 & 16.9 & 56.6 & 60.9
\\
\Xhline{2.0\arrayrulewidth}
\end{tabular}%
}

\caption{Quantitative results of different approaches on Semantic3D (\textit{reduced-8}) \cite{Semantic3D}. This test consists of 78,699,329 points. The scores are obtained from the recent publications. Bold represents the best result in weakly-supervised methods, and underlined represents the best results in fully-supervised methods.}
\label{tab:reduced-8}
\end{table*}


\begin{figure*}[thb]
	\begin{center}
		\includegraphics[width=0.98\textwidth]{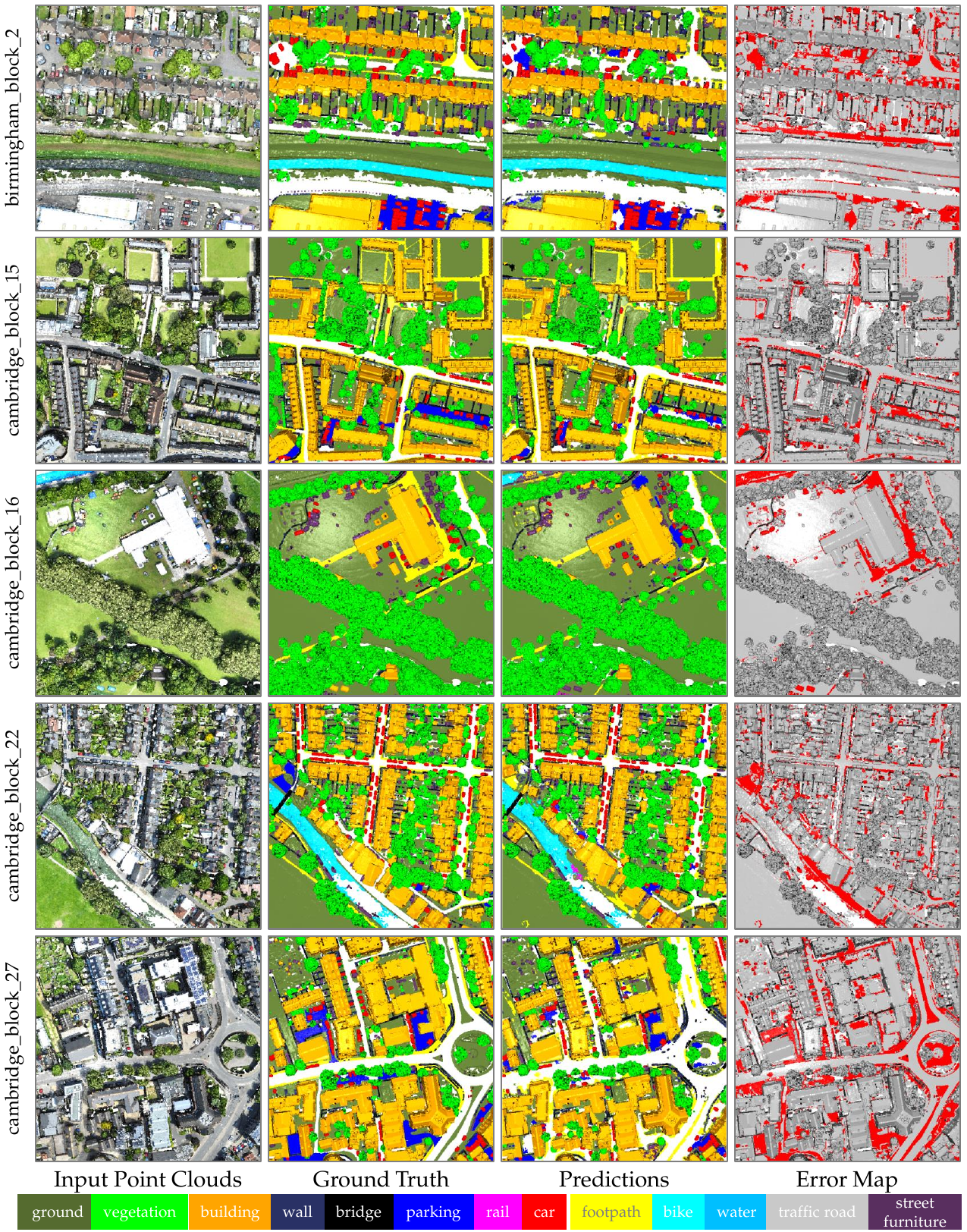}
	\end{center}
    \caption{Qualitative results achieved by our \nickname{} on the validation set (Sequence 08) of SensatUrban\cite{hu2020towards} dataset. Best viewed in color.}
	\label{fig: qualitative_sensaturban}
\end{figure*}

\smallskip\noindent\textbf{(5) Additional Results on SensatUrban.} This is a new urban-scale photogrammetry point cloud dataset covering over 7.6 square kilometers of urban areas in the UK. It has nearly 3 billion points in total. Note that, this dataset is extremely challenging due to the imbalanced class distributions. The detailed experimental results achieved on the SensatUrban dataset are reported in Table \ref{sensaturban}. In addition, we also show the qualitative results achieved by our \nickname{} trained with only 0.1\% labels on this dataset in Fig. \ref{fig: qualitative_sensaturban}. 



\begin{table*}[thb]
\centering
\resizebox{1.0\textwidth}{!}{%
\begin{tabular}{c|rcccccccccccccccc}
\Xhline{2.0\arrayrulewidth}
 Settings & Methods& \rotatebox{90}{OA(\%)} &  \rotatebox{90}{mAcc(\%)} & \rotatebox{90}{\textbf{mIoU(\%)}} & \rotatebox{90}{\textit{ground}} & \rotatebox{90}{\textit{veg.}} & \rotatebox{90}{\textit{building}} & \rotatebox{90}{\textit{wall}} & \rotatebox{90}{\textit{bridge}} & \rotatebox{90}{\textit{parking}} & \rotatebox{90}{\textit{rail}} & \rotatebox{90}{\textit{traffic.}} & \rotatebox{90}{\textit{street.}} & \rotatebox{90}{\textit{car}} & \rotatebox{90}{\textit{footpath}} & \rotatebox{90}{\textit{bike}} & \rotatebox{90}{\textit{water}} \\
\hline
\multirow{7}{*}{\begin{tabular}[c]{@{}c@{}}Full\\ supervision\end{tabular}} 
& PointNet \cite{qi2017pointnet} & 80.78 & 30.32 & 23.71 & 67.96 & 89.52 & 80.05 & 0.00 & 0.00 & 3.95 & 0.00 & 31.55 & 0.00 & 35.14 & 0.00 & 0.00 & 0.00\\
& PointNet++ \cite{qi2017pointnet++} & 84.30 & 39.97 & 32.92 & 72.46 & 94.24 & 84.77 & 2.72 & 2.09 & 25.79 & 0.00 & 31.54 & 11.42 & 38.84 & 7.12 & 0.00 & 56.93\\
& TagentConv \cite{tangentconv} &76.97 & 43.71 & 33.30 & 71.54 & 91.38  & 75.90 & 35.22 & 0.00 & 45.34  & 0.00  &26.69  &19.24  &67.58  &0.01  &0.00  &0.00  \\
& SPGraph \cite{landrieu2018large} & 85.27 & 44.39 & 37.29 & 69.93 & 94.55 & 88.87 & 32.83 & 12.58 & 15.77 & \uline{15.48} & 30.63 & 22.96 & 56.42 & 0.54 & 0.00 & 44.24\\
& SparseConv \cite{sparse} & 88.66 & 63.28 & 42.66 & 74.10 & 97.90 & 94.20 & 63.30 & 7.50 & 24.20 & 0.00 & 30.10 & 34.00 & 74.40 & 0.00 & 0.00 & 54.80\\
& KPConv \cite{thomas2019kpconv} & \uline{93.20} & 63.76 & \uline{57.58} & \uline{87.10} & \uline{98.91} & \uline{95.33} & \uline{74.40} & 28.69 & 41.38 & 0.00 & 55.99 & \uline{54.43} & \uline{85.67} & \uline{40.39} & 0.00 & \uline{86.30}\\
& RandLA-Net \cite{hu2019randla} & 89.78 & \uline{69.64} & 52.69 & 80.11 & 98.07 & 91.58 & 48.88 & \uline{40.75} & \uline{51.62} & 0.00 & \uline{56.67} & 33.23 & 80.14 & 32.63 & 0.00 & 71.31\\
\hline
\multirow{2}{*}{\begin{tabular}[c]{@{}c@{}}Weak\\ supervision\end{tabular}} 
& \textbf{\nicknamenew{} (0.1\%)} & \textbf{90.97} & \textbf{70.84} & \textbf{53.97} & \textbf{83.41} & \textbf{98.22} & \textbf{94.22} & \textbf{48.38} & \textbf{50.84} & \textbf{40.89} & \textbf{14.53} & \textbf{50.72} & \textbf{38.48} & \textbf{75.62} & \textbf{34.03} & 0.00 & \textbf{72.26}  \\
& \textbf{\nicknamenew{} (0.01\%)}  & 85.57 & 49.40 & 37.17 & 74.89 & 96.67 & 88.77 & 32.43 & 7.49 & 12.84 & 0.00 & 29.32 & 22.15 & 67.25 & 0.02 & 0.00 & 51.38 \\
\Xhline{2.0\arrayrulewidth}
\end{tabular}}
\caption{Benchmark results of the baselines on our SensatUrban. Overall Accuracy (OA, \%), mean class Accuracy (mAcc, \%), mean IoU (mIoU, \%), and per-class IoU (\%) scores are reported. Bold represents the best result in weakly-supervised methods, and underlined represents the best results in fully-supervised methods.}
\label{sensaturban}
\end{table*}

\smallskip\noindent\textbf{(6) Additional Results on Toronto3D.} This dataset consists of 1KM urban road point clouds acquired by vehicle-mounted mobile laser systems. It has 78.3 million points belonging to 8 semantic categories. Here, we provide the quantitative comparison of our \nickname{} and several fully-supervised methods in Table \ref{tab:Toronto3D}. Following \cite{hu2019randla}, we also additionally report the performance of our method with and without color information. It can be seen that our \nickname{} outperforms several fully-supervised methods such as KPConv, with merely 0.1\% of point annotations for training. We also notice that the usage of color information closes the gap between our method and the top-performing RandLA-Net \cite{hu2019randla}. This implies that it could be helpful to introduce auxiliary information under the setting of weak supervision.

\begin{table*}[thb]
\centering
\resizebox{0.9\textwidth}{!}{%
\begin{tabular}{c|rcccccccccc}
\Xhline{2.0\arrayrulewidth}
Settings & Methods & OA(\%) & \textbf{mIoU(\%)} & \textit{Road} & \textit{Rd mrk.} & \textit{Natural} & \textit{Building} & \textit{Util. line} & \textit{Pole} & \textit{Car} & \textit{Fence} \\ 
\hline
\multirow{11}{*}{\begin{tabular}[c]{@{}c@{}}Full\\ supervision\end{tabular}} 
& PointNet++ \cite{qi2017pointnet++} & 84.88 & 41.81 & 89.27 & 0.00 & 69.06 & 54.16 & 43.78 & 23.30 & 52.00 & 2.95  \\
& PointNet++ (MSG) \cite{qi2017pointnet++} & 92.56 & 59.47 & 92.90 & 0.00 & 86.13 & 82.15 & 60.96 & 62.81 & 76.41 & 14.43 \\
& DGCNN \cite{dgcnn} & 94.24 & 61.79 & 93.88 & 0.00 & 91.25 & 80.39 & 62.40 & 62.32 & 88.26 & 15.81  \\
& KPFCNN \cite{thomas2019kpconv} & 95.39 & 69.11 & 94.62 & 0.06 & 96.07 & 91.51 & 87.68 & \uline{81.56} & 85.66 & 15.72 \\
& MS-PCNN \cite{ma2019multi} & 90.03 & 65.89 & 93.84 & 3.83 & 93.46 & 82.59 & 67.80 & 71.95 & 91.12 & 22.50 \\
& TGNet \cite{li2019tgnet} & 94.08 & 61.34 & 93.54 & 0.00 & 90.83 & 81.57 & 65.26 & 62.98 & 88.73 & 7.85  \\ 
& MS-TGNet \cite{Toronto3D} & 95.71 & 70.50 & 94.41 & 17.19 & 95.72 & 88.83 & 76.01 & 73.97 & \uline{94.24} & 23.64 \\ 
& RandLA-Net (w/ RGB)$^\dagger$ \cite{hu2019randla} & \uline{97.15} & \uline{81.88} & \uline{96.69} & \uline{64.10} & \uline{96.85} & \uline{94.14} & \uline{88.03} & 77.48 & 93.21 & \uline{44.53} \\
& RandLA-Net (w/o RGB) \cite{hu2019randla} & 95.63 & 77.72 & 94.53 & 42.44 & 96.62 & 93.10 & 86.56 & 76.83 & 92.55 & 39.14 \\ 
\hline
\multirow{2}{*}{\begin{tabular}[c]{@{}c@{}}Weak\\ supervision\end{tabular}} 
& \textbf{\nicknamenew{} (w/ RGB, 0.1\%)}$^\dagger$ & \textbf{96.67} & \textbf{77.75} & \textbf{96.69} & \textbf{65.67} & \textbf{94.58} & \textbf{91.34} & \textbf{83.36} & \textbf{70.59} & \textbf{88.87} & \textbf{30.91}  \\
& \textbf{\nicknamenew{} (w/o RGB, 0.1\%)} & 92.84 & 69.35 & 93.74 & 16.83 & 92.55 & 89.04 & 82.50 & 63.98 & 88.17 & 28.01 \\
& \textbf{\nicknamenew{} (w/ RGB, 0.01\%)}$^\dagger$ &94.19 & 68.17 & 95.26 & 54.44 & 88.20 & 84.07 & 75.87 & 57.52 & 84.33 & 5.69 \\
& \textbf{\nicknamenew{} (w/o RGB, 0.01\%)} & 90.47 & 57.57 & 90.97 & 4.99 & 84.10 & 80.29 & 62.78 & 56.51 & 69.49 & 11.44 \\
\Xhline{2.0\arrayrulewidth}
\end{tabular}
}
\caption{Quantitative results of different approaches on the Toronto3D \cite{Toronto3D} dataset. The scores of the baselines are obtained from \cite{Toronto3D}.  Bold represents the best result in weakly-supervised methods, and underlined represents the best results in fully-supervised methods.}
\label{tab:Toronto3D}
\end{table*}

\smallskip\noindent\textbf{(7) Additional Results on DALES.} This dataset consists of large-scale earth scans acquired by an aerial LiDAR. It covers over 10 $km^2$ spatial ranges with 5 million points belonging to 8 semantic categories. We compare our \nickname{} with strong fully-supervised approaches. As shown in Table \ref{tab:dales}, our method achieves higher mIoU scores than PointNet++ \cite{qi2017pointnet++}, ConvPoint \cite{boulch2020lightconvpoint}, SPGraph \cite{landrieu2018large}, PointCNN \cite{li2018pointcnn} and ShellNet \cite{zhang2019shellnet}, with only 0.1\% labels for training. However, there is still a performance gap compared with the leading fully-supervised counterparts such as RandLA-Net \cite{hu2019randla}, primarily due to our weak performance on minor categories such as \textit{trucks} and \textit{cars}. The potential reason is that the simple random annotation strategy may happen to ignore the underrepresented classes.


\begin{table*}[thb]
\begin{center}
\resizebox{0.9\textwidth}{!}{%
\begin{tabular}{c|rcccccccccc}
\Xhline{2.0\arrayrulewidth}
Settings & Method & OA(\%) & \textbf{mIoU(\%)} & \textit{ground}& \textit{buildings}&\textit{cars}&\textit{trucks}&\textit{poles}&\textit{power lines}&\textit{fences}&\textit{vegetation}      \\
\hline
\multirow{7}{*}{\begin{tabular}[c]{@{}c@{}}Full\\ supervision\end{tabular}} 
& ShellNet \cite{zhang2019shellnet} & 96.4&	57.4& 96.0 &	95.4&	32.2&	39.6&	20.0&	27.4&	60.0&	88.4\\
& PointCNN \cite{li2018pointcnn} & 97.2	&58.4&	97.5 &	95.7&	40.6&	4.80&	57.6&	26.7&	52.6&	91.7\\
& SuperPoint \cite{landrieu2018large} &95.5&	60.6&	94.7&	93.4&	62.9&	18.7&	28.5&	65.2&33.6&	87.9\\
& ConvPoint \cite{conv_pts} &97.2&	67.4&	96.9&	96.3&	75.5&	21.7&	40.3&	86.7&	29.6&	91.9\\
& PointNet++ \cite{qi2017pointnet++} & 95.7	&68.3	&94.1&	89.1&	75.4&	30.3&	40.0& 79.9&	46.2&	91.2\\
& KPConv \cite{thomas2019kpconv}& 97.8&	81.1&	97.1&	96.6&	85.3&	41.9&	75.0&	95.5&	63.5&	94.1\\
& RandLA-Net \cite{hu2019randla} & 97.1	&80.0	&97.0&	93.2&	83.7&	43.8&	59.4& 94.8&	\uline{71.5} &	\uline{96.6} \\
& Pyramid Point \cite{varney2020pyramid} & \uline{98.3} & \uline{83.6} & \uline{97.8} & \uline{97.3} & \uline{88.4} & \uline{47.9} & \uline{77.6} & \uline{96.7} & 67.5 & 95.4 \\
\hline
\multirow{2}{*}{\begin{tabular}[c]{@{}c@{}}Weak\\ supervision\end{tabular}} 
& \textbf{\nicknamenew{} (0.1\%)} & \textbf{97.1} & \textbf{72.0} & \textbf{96.7} & \textbf{92.0} & \textbf{75.2} & \textbf{27.3} & \textbf{87.4} & \textbf{48.1} & \textbf{53.7} & \textbf{95.8} \\
& \textbf{\nicknamenew{} (0.01\%)}  &95.9  & 60.4 & 95.9 & 90.1 & 57.7 & 12.8 & 75.2 & 32.9 & 24.9 & 93.4\\
\Xhline{2.0\arrayrulewidth}
\end{tabular}}
\end{center}
\caption{Quantitative results of different approaches on the DALES dataset. Overall Accuracy (OA, \%), mean class Accuracy (mAcc, \%), mean IoU (mIoU, \%), and per-class IoU (\%) are reported. Bold represents the best result in weakly-supervised methods, and underlined represents the best results in fully-supervised methods.}
\label{tab:dales}
\end{table*}

\begin{figure*}[thb]
	\begin{center}
		\includegraphics[width=1.0\textwidth]{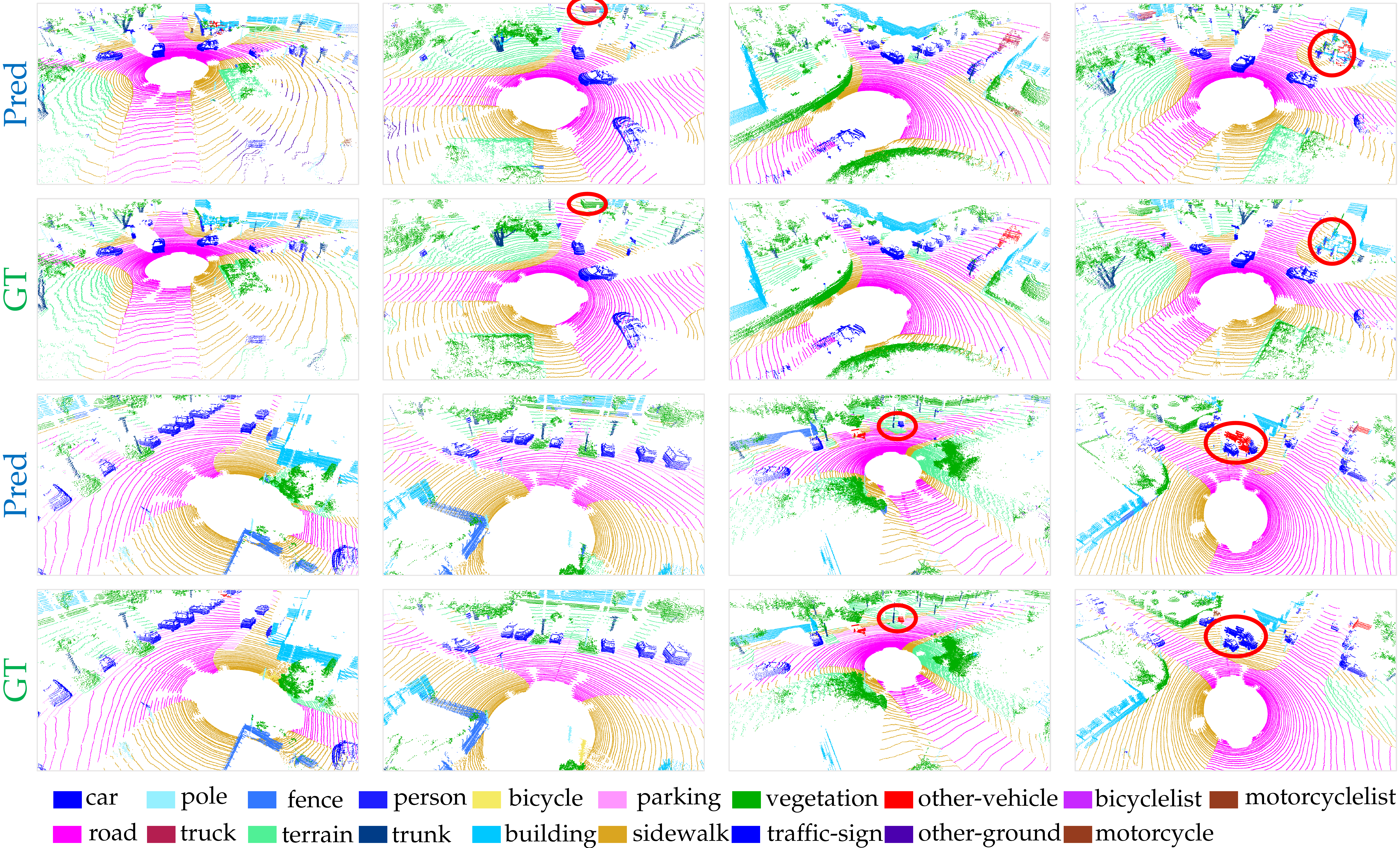}
	\end{center}
    \caption{Qualitative results achieved by our \nickname{} on the validation set (Sequence 08) of SemanticKITTI \cite{behley2019semantickitti} dataset. The red circle highlights the failure case.}
	\label{fig: qualitative_kitti}
\end{figure*}

\smallskip\noindent\textbf{(8) Additional Results on SemanticKITTI.} This large-scale dataset consists of point cloud sequences captured by LiDAR for autonomous driving. In particular, it has 22 sequences, 43552 sparse scans, and nearly 4 billion points. Note that, RGB is not available in this dataset. We compare our \nickname{} with fully-supervised techniques on the online test set in Table \ref{tab:SemanticKITTI}. It can be seen that our approach achieves a satisfactory mIoU score, outperforming several strong baselines with only 0.1\% labels for training. In addition, our model only has 1.05 million trainable parameters, and is extremely lightweight and suitable for real-world applications. Finally, we also visualize the segmentation results achieved by our \nickname{} on the validation set of the SemanticKITTI dataset in Fig. \ref{fig: qualitative_kitti}.


\begin{table*}[thb]
\centering
\resizebox{\textwidth}{!}{%
\begin{tabular}{c|rccccccccccccccccccccc}
\Xhline{2.0\arrayrulewidth}
Settings & Methods & \rotatebox{90}{\textbf{mIoU(\%)}} & \rotatebox{90}{Params(M)} & \rotatebox{90}{\textit{road}} & \rotatebox{90}{\textit{sidewalk}} & \rotatebox{90}{\textit{parking}} & \rotatebox{90}{\textit{other-ground}} & \rotatebox{90}{\textit{building}} & \rotatebox{90}{\textit{car}} & \rotatebox{90}{\textit{truck}} & \rotatebox{90}{\textit{bicycle}} & \rotatebox{90}{\textit{motorcycle}} & \rotatebox{90}{\textit{other-vehicle}} & \rotatebox{90}{\textit{vegetation}} & \rotatebox{90}{\textit{trunk}} & \rotatebox{90}{\textit{terrain}} & \rotatebox{90}{\textit{person}} & \rotatebox{90}{\textit{bicyclist}} & \rotatebox{90}{\textit{motorcyclist}} & \rotatebox{90}{\textit{fence}} & \rotatebox{90}{\textit{pole}} & \rotatebox{90}{\textit{traffic-sign}} \\
\hline
\multirow{15}{*}{\begin{tabular}[c]{@{}c@{}}Full\\ supervision\end{tabular}} 
& PointNet \cite{qi2017pointnet} & 14.6 & 3  & 61.6 & 35.7 & 15.8 & 1.4 & 41.4 & 46.3 & 0.1 & 1.3 & 0.3 & 0.8 & 31.0 & 4.6 & 17.6 & 0.2 & 0.2 & 0.0 & 12.9 & 2.4 & 3.7 \\
& SPG \cite{landrieu2018large}  & 17.4 & \uline{0.25}  & 45.0 & 28.5 & 0.6 & 0.6 & 64.3 & 49.3 & 0.1 & 0.2 & 0.2 & 0.8 & 48.9 & 27.2 & 24.6 & 0.3 & 2.7 & 0.1 & 20.8 & 15.9 & 0.8 \\
& SPLATNet \cite{su2018splatnet}  & 18.4 & 0.8  & 64.6 & 39.1 & 0.4 & 0.0 & 58.3 & 58.2 & 0.0 & 0.0 & 0.0 & 0.0 & 71.1 & 9.9 & 19.3 & 0.0 & 0.0 & 0.0 & 23.1 & 5.6 & 0.0 \\
& PointNet++ \cite{qi2017pointnet++}  & 20.1 & 6 & 72.0 & 41.8 & 18.7 & 5.6 & 62.3 & 53.7 & 0.9 & 1.9 & 0.2 & 0.2 & 46.5 & 13.8 & 30.0 & 0.9 & 1.0 & 0.0 & 16.9 & 6.0 & 8.9 \\
& TangentConv \cite{tangentconv} & 40.9 & 0.4 & 83.9 & 63.9 & 33.4 & 15.4 & 83.4 & 90.8 & 15.2 & 2.7 & 16.5 & 12.1 & 79.5 & 49.3 & 58.1 & 23.0 & 28.4 & 8.1 & 49.0 & 35.8 & 28.5 \\
& LatticeNet \cite{rosu2019latticenet} & 52.2 & - & 88.8 & 73.8 & 64.6 & 25.6 & 86.9 & 88.6 & 43.3 & 12.0 & 20.8 & 24.8 & 76.4 & 57.9 & 54.7 & 34.2 & 39.9 & \uline{60.9} & 55.2 & 41.5 & 42.7\\
& PolarNet \cite{zhang2020polarnet} & 54.3 & 14  & 90.8 & 74.4 &61.7 &21.7 &\uline{90.0} &93.8 & 22.9 &40.2 & 30.1 &28.5 &\uline{84.0} &\uline{65.5} &67.8 &43.2 & 40.2 &5.6 &\uline{61.3} & \uline{51.8} & \uline{57.5}\\
& RandLA-Net \cite{hu2019randla} & \uline{55.9} & 1.24  & 90.5 & 74.0 & 61.8 & 24.5 & 89.7 & \uline{94.2} & \uline{43.9} & \uline{47.4} & 32.2 & \uline{39.1} & 83.8 & 63.6 & \uline{68.6} & 48.4 & 47.4 & 9.4 & 60.4 & 51.0 & 50.7 \\ \cline{2-23}
& SqueezeSeg \cite{wu2018squeezeseg}  & 29.5 & 1 & 85.4 & 54.3 & 26.9 & 4.5 & 57.4 & 68.8 & 3.3 & 16.0 & 4.1 & 3.6 & 60.0 & 24.3 & 53.7 & 12.9 & 13.1 & 0.9 & 29.0 & 17.5 & 24.5 \\
& SqueezeSegV2 \cite{wu2019squeezesegv2}  & 39.7 & 1  & 88.6 & 67.6 & 45.8 & 17.7 & 73.7 & 81.8 & 13.4 & 18.5 & 17.9 & 14.0 & 71.8 & 35.8 & 60.2 & 20.1 & 25.1 & 3.9 & 41.1 & 20.2 & 36.3 \\
& DarkNet21Seg \cite{behley2019semantickitti}  & 47.4 & 25  & 91.4 & 74.0 & 57.0 & 26.4 & 81.9 & 85.4 & 18.6 & 26.2 & 26.5 & 15.6 & 77.6 & 48.4 & 63.6 & 31.8 & 33.6 & 4.0 & 52.3 & 36.0 & 50.0 \\
& DarkNet53Seg \cite{behley2019semantickitti} & 49.9 & 50 & \uline{91.8} & 74.6 & 64.8 & \uline{27.9} & 84.1 & 86.4 & 25.5 & 24.5 & 32.7 & 22.6 & 78.3 & 50.1 & 64.0 & 36.2 & 33.6 & 4.7 & 55.0 & 38.9 & 52.2 \\
& RangeNet53++ \cite{rangenet++} &  52.2 & 50  & \uline{91.8} & \uline{75.2} & \uline{65.0} & 27.8 & 87.4 & 91.4 & 25.7 & 25.7 & \uline{34.4} & 23.0 & 80.5 & 55.1 & 64.6 &38.3  & 38.8  & 4.8 & 58.6 & 47.9 & \uline{55.9} \\
& SalsaNext \cite{salsanext} &  54.5 & 6.73 & 90.9 & 74.0 & 58.1 & 27.8 & 87.9 & 90.9 & 21.7 & 36.4 & 29.5 & 19.9 & 81.8 & 61.7 & 66.3 & \uline{52.0} & \uline{52.7} & 16.0 & 58.2 & 51.7 & 58.0 \\
& SqueezeSegV3 \cite{xu2020squeezesegv3}  &\uline{55.9} &26  & 91.7 & 74.8 & 63.4 & 26.4 & 89.0 & 92.5 & 29.6 & 38.7 & 36.5 & 33.0 & 82.0 & 58.7 & 65.4 & 45.6 & 46.2 & 20.1 & 59.4 & 49.6 & 58.9 \\
\hline
\multirow{2}{*}{\begin{tabular}[c]{@{}c@{}}Weak\\ supervision\end{tabular}} 
& \textbf{\nicknamenew{} (0.1\%)}  & \textbf{50.8} & 1.05 & \textbf{90.5} & \textbf{72.9} & \textbf{56.8} & \textbf{19.1} & \textbf{84.8} & \textbf{92.1} & \textbf{36.7} & \textbf{39.3} & \textbf{30.1} & \textbf{26.0} & \textbf{80.8} & \textbf{59.1} & \textbf{67.0} & \textbf{36.4} & \textbf{25.3} & \textbf{7.2} & \textbf{53.3} & \textbf{44.5} & \textbf{44.0}     \\
& \textbf{\nicknamenew{} (0.01\%)} & 39.1 & 1.05 & 86.6 & 66.4 & 43.0 & 16.9 & 80.0 & 85.5 & 12.9 & 4.0 & 1.4 & 18.4 & 72.7 & 49.6 & 58.8 & 16.9 & 22.3 & 4.3 & 42.3 & 31.7 & 16.6 \\
\Xhline{2.0\arrayrulewidth}
\end{tabular}%
}
\caption{Quantitative results of different approaches on SemanticKITTI \cite{behley2019semantickitti}. The scores are obtained from the recent publications.  Bold represents the best result in weakly-supervised methods, and underlined represents the best results in fully-supervised methods.}
\label{tab:SemanticKITTI}
\end{table*}

\end{document}